\documentclass{article}

\usepackage[main, final]{neurips_2026}

\usepackage[utf8]{inputenc} 
\usepackage[T1]{fontenc}    
\usepackage{hyperref}       
\usepackage{url}            
\usepackage{booktabs}       
\usepackage{amsfonts}       
\usepackage{nicefrac}       
\usepackage{microtype}      
\usepackage{xcolor}         
\usepackage{graphicx}       
\usepackage{amsmath}
\usepackage{subcaption}     
\usepackage{multirow}
\usepackage{multicol}
\usepackage{tabularx}
\usepackage{caption}
\usepackage{wrapfig}
\usepackage{float}
\usepackage[most]{tcolorbox}
\usepackage{enumitem}
\usepackage{svg}

\title{Aligning MLLMs with the Latent Structure of Human Cognition via Behavior-Derived Semantic Dimensions}

\author{
  \textbf{Ning E}$^{1,2}$\thanks{Equal contribution.} \quad
  \textbf{Changde Du}$^{2,5}$\footnotemark[1] \quad
  \textbf{Yizhuo Lu}$^{2,3}$\footnotemark[1] \quad
  \textbf{Huiguang He}$^{2,3,4,1,5}$\thanks{Corresponding author: huiguang.he@ia.ac.cn.} \\[2ex]
  $^1$School of Biomedical Engineering, ShanghaiTech University, Shanghai, China  \\
  $^2$Institute of Automation, Chinese Academy of Sciences, Beijing, China \\
  $^3$School of Future Technology, University of Chinese Academy of Sciences, Beijing, China \\
  $^4$School of Artificial Intelligence, University of Chinese Academy of Sciences, Beijing, China \\
  $^5$Zhongguancun Academy, Beijing, China
}

\begin{document}

\maketitle

\begin{abstract}
  Multimodal Large Language Models (MLLMs) have achieved impressive performance across a wide range of reasoning tasks, yet it remains unclear whether their internal representations and decision criteria align with the latent structure of human cognition. In this work, we propose a novel framework to align MLLMs with the multidimensional mental representations underlying human similarity judgements. We build on a previously established behavior-derived semantic embedding space, in which objects are represented along 66 interpretable dimensions derived from 4.70 million human Odd-One-Out (O1O) judgements. For each triplet, we infer the most salient decision dimension and convert it into a dimension-grounded linguistic rationale. This enables MLLMs to learn both the final choice and a justification consistent with the inferred behavioral criterion. Experimental results demonstrate that our approach improves model consistency with human similarity judgements while retaining competitive performance on standard multimodal benchmarks. Furthermore, using searchlight Representational Similarity Analysis (RSA) on an independent fMRI Natural Scenes Dataset (NSD), we observe increased representational alignment between model and human. Ultimately, our findings point to a data-driven route for incorporating human cognitive structure into MLLMs alignment. The code is available \href{https://github.com/ening-open/cogalign}{here}.
\end{abstract}

\section{Introduction}
\label{sec:intro}

The advent of Multimodal Large Language Models (MLLMs) marks a major milestone in artificial intelligence, demonstrating strong performance across a spectrum of cognitive and reasoning tasks \citep{qwen35blog,qwen3vl,glm5} and naturally emerging human-like representations \citep{duhumanlike}. Current alignment paradigms \citep{dreamsim, dreamsim2,instructgpt,ppo,dpo} focus on outcome-level behavioral mimicry, optimizing models to generate outputs that match human preferences without deeper human cognitive logic. Relying on this shallow mimicry leaves it unclear whether models solve tasks using human-consistent decision criteria or by exploiting statistical shortcuts. To realize more robust generalization, intrinsic interpretability, and authentic anthropomorphism, MLLMs alignment requires moving beyond behavioral mimicry toward the latent structure of human cognition.

To probe this alignment gap, we utilize the Odd-One-Out (O1O) task, a cornerstone of cognitive psychology for evaluating conceptual representations and reasoning. In this task, a subject is presented with three objects and must identify the one that is least similar to the other two. Identifying the odd object requires accessing the latent semantic dimensions that humans use to structure their conceptual world. Evaluating MLLMs on this task demonstrates that, despite their robust general capabilities, a persistent performance gap remains in predicting human choices. This disparity reveals a misalignment: standard approaches primarily align models with raw behavioral outcomes (i.e., what humans choose), failing to capture the underlying human cognition that dictates the decision criteria (i.e., why they choose it).

Our key insight to bridge this gap is to transform the latent structure of human cognition into explicit and learnable supervision. To achieve this, we build on the previous work of Hebart et al. \citep{hebart2020revealing,hebart2023things}, which established behavior-derived semantic embedding space. By analyzing 4.70 million human O1O judgements, they reveal interpretable 66-dimensional mental representations of natural objects underlying human similarity judgements. We adopt the 66-dimensional mental representations as the latent structure of human cognition to align MLLMs, enabling MLLMs to learn not only what humans choose, but also why they make choice. Concretely, for each triplet, we identify the most influential dimension driving human decision-making via the Jackknife \citep {mahner2025dimensions} method. However, a discrete semantic dimension is insufficient to bridge the gap with generative modality of MLLMs. To address this, we convert the discrete dimension into a dimension-grounded linguistic rationale, which serves as an explicit Chain-of-Thought (CoT) constraint.

Building on this formulation, we construct the alignment dataset to fine-tune MLLMs, thereby deeply aligning their reasoning processes with the latent structure of human cognition. This is the concrete implementation of our framework. To validate the efficacy of our framework, we evaluate our framework along multiple axes. Behaviorally, the aligned MLLMs improve consistency with human similarity judgements compared to baselines, and retain competitive performance on standard multimodal benchmarks. Beyond behavioral aspect, we conduct searchlight Representational Similarity Analysis (RSA) \citep{rsa} using the independent fMRI Natural Scenes Dataset (NSD) \citep{allen2022massive}. The RSA results suggest that the internal representations of our aligned MLLMs demonstrate improved representational similarity to human neural activity. These neural results are not to reveal biological localization, but as a convergent, independent measure of similarity. Our findings demonstrate that explicitly incorporating behavior-derived cognitive priors is an effective and viable pathway for advancing the representational alignment between MLLMs and human cognitive structure.

Our contributions are summarized as follows: (1) We propose a novel alignment framework that moves beyond shallow behavioral mimicry by explicitly internalizing the latent structure of human cognition. By converting behavior-derived semantic dimensions into dimension-grounded linguistic rationales, MLLMs learn human-like decision-making cognition. (2) We introduce a pathway to integrate multidimensional mental representations from cognitive science into MLLMs. (3) Guided by this framework, we construct the alignment dataset that augments human similarity judgements with explicit rationales.
\section{Related Work}
\label{sec:related-work}

Recent advances have made significant strides in aligning models with human cognitive behavior. Muttenthaler et al. improve neural network representations using human similarity judgements \citep{muttenthaler2023improving} and find that model scale and architecture have essentially no effect on the alignment with human behavioral responses, whereas the training dataset and objective function both have a much larger impact \citep{muttenthaler2023hannr}. AligNet \citep{muttenthaler2025aligningma} distills human similarity judgements from the THINGS dataset \citep{hebart2023things} into vision models, successfully embedding human-like conceptual hierarchies into visual representations. In the video domain, Garcia and Isik \citep{garcia2026bgs} introduce behavioral geometric supervision to align video representations with human judgements of social similarity, improving their correspondence with human social perception. Similarly, in the domain of Large Language Models (LLMs), Centaur \citep{binz2025foundation} fine-tunes LLMs on an unprecedented scale of human behavioral data (Psych-101), demonstrating remarkable capabilities in predicting and simulating human behavior in any experiment expressible in natural language.

While these pioneering works establish the viability of using human behavioral data for model alignment, they primarily operate under the paradigm of outcome-level behavioral mimicry. By optimizing models directly toward human decision outcomes, these approaches implicitly assume that matching the output distribution entails capturing the underlying cognitive process. However, this leaves a critical gap: it remains opaque whether the models genuinely internalize the latent cognitive structure driving these decisions.

Our work builds on the foundation laid by these efforts but takes a step forward. Rather than relying on outcome matching, we argue that cognitive alignment in MLLMs requires explicit supervision of the reasoning process. By leveraging established behavior-derived semantic dimensions to construct dimension-grounded linguistic rationales, our framework explicitly models why a choice is made. This transforms the implicit cognitive structure into learnable CoT constraints, advancing the alignment paradigm from shallow behavioral mimicry to the integration of latent human cognitive structure.

\section{Method}
\label{sec:method}

\subsection{Triplet-Level Semantic Dimension Attribution}
\label{method:dim-attr}
\begin{figure}[!htbp]
  \centering
  \includegraphics[width=0.88\linewidth]{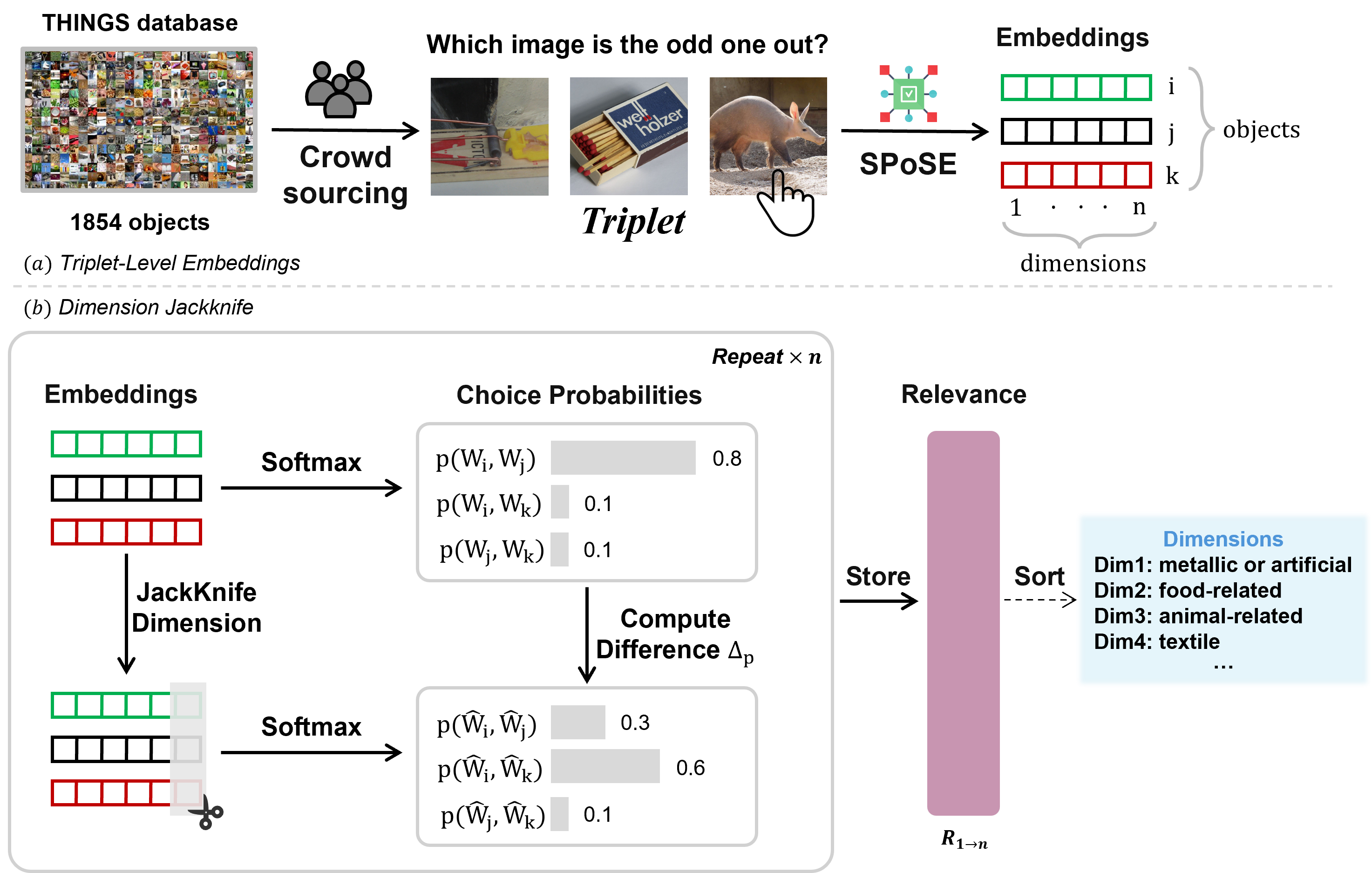}
  \caption{Overview of triplet-level semantic dimension attribution. (a) Human similarity judgements are collected via crowdsourcing. Then learning a representational embedding space via SPoSE yields the multidimensional mental representations of natural objects. (b) For one triplet, we compute the original softmax probability based on the entire representational embedding for all pairs. We then iteratively prune each dimension from the representational embedding space and compute the changes in softmax probabilities relative to those obtained with the full embedding as the relevance score for that dimension. By sorting these relevance scores in descending order, we establish a hierarchical ranking of dimensional importance. The \textbf{top-1} dimension yielding the maximum probability change is regarded as the most salient semantic dimension driving the human decision-making.}
  \label{fig:dim-attr}
\end{figure}

As illustrated in Figure~\ref{fig:dim-attr}, we first build on the established behavior-derived semantic embedding space introduced by Hebart et al. \citep{hebart2020revealing,hebart2023things}, specifically adopting the object-level semantic dimensions estimated via the Sparse Positive Similarity Embedding (SPoSE) \citep{spose} model. To bridge these object representations with each triplet decision, we adopt a Jackknife-based attribution method to identify the most influential dimension underlying human similarity judgement.

\paragraph{Object embedding space.}
As stated in the preliminaries~\ref{sec:prel}, given triplet objects $\mathcal{T} = \{i, j, k\}$ ($k$ is odd-one-out), we define the probability that any two objects $\{u, v\} \subset \mathcal{T}$ are identified as the most similar pair as $p(W_u, W_v)$. This probability is computed via a softmax over the inner products of their embeddings $W_u$ and $W_v$.

\paragraph{Triplet-level dimensions jackknife.}
To infer which specific semantic dimension $d^*$ primarily drives a human decision (top-1 dimension), we employ a dimension Jackknife method. We iteratively ablate individual dimensions from the triplet objects embeddings. Let $\hat{W}$ be the embedding vector where the $d$ dimension is pruned. We then recompute the ablated choice probability, denoted as $p(\hat{W}_u, \hat{W}_v)$. The relevance score of dimension $d$ to the human judgement is quantified by the absolute probability difference:
\begin{equation}
  \Delta_p = \left| p(W_i, W_j) - p(\hat{W}_i, \hat{W}_j) \right|
\end{equation}
A larger $\Delta_p$ indicates that the original human choice is more sensitive to that pruned dimension. This process is repeated for all $n=66$ dimensions to yield a relevance scores vector $R_{1 \dots 66}$.

\subsection{Dimension-Guided Reasoning Enhancement}
\label{method:rationale-generation}
While the Jackknife attribution effectively isolates the most influential semantic dimension $d^{*}$ driving human decision-making, utilizing discrete semantic labels (e.g., metallic or artificial) provides a sparse and contextually limited supervision signal. To address this, we convert the discrete semantic dimension $d^{*}$ label into an explicit, dimension-guided linguistic rationale $r(d^{*})$ leveraging the generation capabilities of LLM.

\begin{figure}[h]
  \centering
  \includegraphics[width=0.96\textwidth]{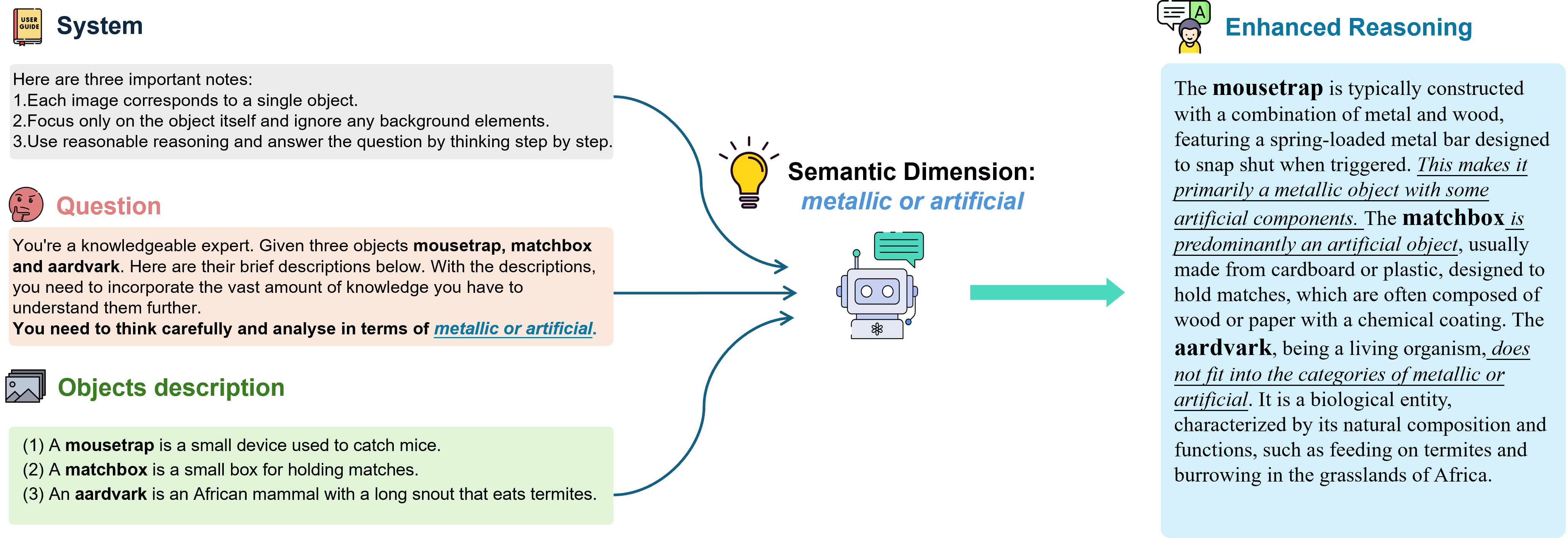}
  \caption{Pipeline for dimension-grounded reasoning enhancement. This pipeline illustrates the process of reasoning enhancement via LLM. For one triplet with their textual descriptions and the most salient semantic dimension $d^*$ driving human decision-making, we integrate them into the structured prompt $Prompt(\mathcal{T}, cap(\mathcal{T}), d^*)$. LLM is then utilized exclusively to enhance reasoning. By injecting the dimension $d^*$, we guide the LLM to perform targeted reasoning enhancement. This effectively converts discrete cognitive criteria for decision-making into explicit CoT rationale $r(d^*)$.}
  \label{fig:rationale-generation}
\end{figure}

As shown in Figure~\ref{fig:rationale-generation}, for one triplet $\mathcal{T}$ and corresponding textual descriptions $cap(\mathcal{T})$, we synthesize the structured prompt~\ref{prompt:rationale-generation}, $Prompt(\mathcal{T}, cap(\mathcal{T}), d^{*})$. Crucially, the LLM is not tasked with predicting the human choice. Instead, it is utilized exclusively as a reasoning enhancement engine. By injecting the semantic dimension $d^{*}$ into the prompt context, we guide the LLM to articulate exactly how this dimension drives human thinking and decision-making. The LLM processes the prompt to synthesize a coherent CoT reasoning path $r(d^{*})$. Ultimately, these dimension-guided rationales provide high-quality, cognitively aligned supervisory data for the MLLMs alignment.

\subsection{Cognitive Latent Structure Alignment}
The final stage of our framework is to align MLLMs via Supervised Fine-Tuning (SFT) using the derived explicit linguistic rationale $r(d^*)$. Current alignment paradigms predominantly rely on outcome-level supervision, optimizing models for behavioral mimicry without internalizing the underlying latent structure of human cognition. By explicitly anchoring the MLLMs in human-guided reasoning trajectories, we steer their internal decision-making mechanisms to align with the latent structure of human cognition.

\paragraph{Alignment optimization objective.}
Given the visual input of triplet objects $v$ and the text instruction $x_{\text{inst}}$, the enhanced reasoning trajectory is denoted as $\tau = [r(d^*); y]$. This formulation naturally factorizes the generative probability into two sequential cognitive phases: rationale generation (thinking) and final decision-making. Let $\theta$ parameterize the MLLMs; the alignment objective over our dataset $\mathcal{D}_{\text{O1O}}$ is formulated as:

\begin{center}
  \small
  \begin{equation}
    \mathcal{L}_{\text{O1O}}(\theta) = - \mathbb{E}_{(\mathbf{v}, x_{\text{inst}}, \tau) \in \mathcal{D}_{\text{O1O}}} \left[ \sum_{t=1}^{\tau_r} \log P_\theta(r_t \mid \mathbf{v}, x_{\text{inst}}, r_{<t}) + \sum_{j=\tau_{r}+1}^{\tau_r+\tau_y} \log P_\theta(y_j \mid \mathbf{v}, x_{\text{inst}},\tau_r, y_{<j}) \right]
  \end{equation}
\end{center}

where $P_\theta(\cdot)$ denotes the autoregressive next-token prediction probability parameterized by the MLLMs. In this formulation, $\tau_r$ and $\tau_y$ denote the sequence lengths of the rationale and the final decision-making, respectively. During the initial reasoning phase ($t \in [1, \tau_r]$), $r_t$ represents the $t$-th rationale token generated conditioned on the visual inputs $\mathbf{v}$, the text instruction $x_{\text{inst}}$, and the preceding tokens $r_{<t}$. In the subsequent decision-making phase ($j \in [\tau_r+1, \tau_r+\tau_y]$), $y_j$ denotes the $j$-th target token. Crucially, this second phase is conditioned on $\tau_r$, which herein represents the fully generated rationale sequence acting as a cognitive anchor. By optimizing this negative log-likelihood, we constrain the MLLMs internal decision-making mechanisms to internalize the latent human-like structure before the final decision.
\paragraph{Joint optimization.}
While $\mathcal{L}_{\text{O1O}}(\theta)$ effectively instills cognitive latent structure, fine-tuning exclusively on a narrow task distribution risks Catastrophic Forgetting (CF). To preserve foundational capabilities while aligning, we employ a joint optimization strategy. We use multi-task objective by mixing $\mathcal{D}_{\text{O1O}}$ with a general-domain dataset $\mathcal{D}_{\text{gen}}$ (e.g., visual question answering data):
\begin{equation}
  \mathcal{L}(\theta) = \mathcal{L}_{\text{O1O}}(\theta) + \lambda \mathcal{L}_{\text{gen}}(\theta)
\end{equation}

where $\mathcal{L}_{\text{gen}}(\theta)$ denotes the standard autoregressive next-token prediction loss on general multimodal data, ensuring the model retains its foundational generative capabilities. $\lambda$ is a regularization hyperparameter, implemented via dynamic data sampling ratios during training (see Appendix~\ref{apx:data-sampling-strategy} for details). This multi-objective formulation allows the MLLMs to internalize human-consistent reasoning while retaining their broad generalization.

\subsection{Neural Searchlight RSA}
\label{method:neural-rsa}
To provide a convergent, independent measure of representational alignment beyond behavioral evaluations, we conduct searchlight RSA using NSD. This neural evaluation is not intended to claim precise biological localization of MLLMs features, but rather to serve as a robust, modality-independent validation of our alignment framework. Given the same visual stimuli $\mathcal{S}$ for both MLLMs and human, we denote this as $\mathcal{S} = \{s_1, s_2 \dots, s_n\}$ and compute Representational Dissimilarity Matrix (RDM).

\paragraph{MLLMs RDM.}
After extracting the internal representations via a forward propagation (detailed in Appendix~\ref{apx:feature-extraction-method}), we denote $h_i \in \mathbb{R}^D$ as the $D$-dimensional hidden state vector extracted from the \texttt{<|vision\_end|>} token for $s_i$. We quantify the internal representational geometry by computing the MLLMs RDM, denoted as $RDM^{M} \in \mathbb{R}^{S \times S}$. The pairwise representational dissimilarity between any two stimuli $s_i$ and $s_j$ is defined as:
\begin{equation}
  RDM^{M}_{i,j} = 1 - \rho(h_i, h_j)
\end{equation}

where $\rho(\cdot, \cdot)$ is the Pearson correlation coefficient across the $D$ feature dimensions.

\paragraph{Human RDM.}
For a spherical searchlight neighborhood centered at voxel $c$ comprising $C$ contiguous voxels (radius = 3 voxels), let $b_i^{c} \in \mathbb{R}^C$ denote the local multi-voxel neural activation pattern elicited by stimulus $s_i$. The corresponding localized RDM, denoted as $RDM^{c} \in \mathbb{R}^{S \times S}$, is computed by:
\begin{equation}
  RDM^{c}_{i,j} = 1 - \rho(b_i^{c}, b_j^{c})
\end{equation}

\paragraph{Searchlight RSA score.}
To quantify the representational alignment between the MLLMs internal representation space and the localized human cortex, we compute the voxel-wise neural alignment score, $RSA(\mathcal{M}, c)$. Let $u(\cdot)$ denote the operator that extracts the flattened upper triangular elements of a given RDM, excluding the main diagonal. The alignment score at searchlight center $c$ is defined as the Spearman correlation coefficient between the respective upper triangular vectors:
\begin{equation}
  RSA(\mathcal{M}, c) = \rho_s(u(RDM^{M}), u(RDM^{c}))
\end{equation}
To evaluate the representational enhancement driven by our alignment framework, we compute the voxel-wise alignment shift, denoted as $\Delta_{RSA}(c)$. This shift quantifies the difference in alignment scores between the aligned MLLMs ($\mathcal{M}_{align}$) and the unaligned baseline ($\mathcal{M}_{base}$):
\begin{equation}
  \Delta_{RSA}(c) = RSA(\mathcal{M}_{align}, c) - RSA(\mathcal{M}_{base}, c)
\end{equation}

Prior to cortical projection, we apply a lower-bound threshold to the alignment shift map to filter out negligible negative fluctuations. Empirically, negative $\Delta_{RSA}(c)$ values are extremely close to zero and arise from computational noise rather than meaningful representational degradation. Therefore, we clamp these values to a near-zero floor ($\epsilon = 10^{-8}$):
\begin{equation}
  \tilde{\Delta}_{RSA}(c) = \max(\epsilon, \Delta_{RSA}(c))
\end{equation}
The resulting non-negative alignment shift map, $\tilde{\Delta}_{RSA}$, is then projected onto a flattened cortical surface using pycortex \citep{gao2015pycortex}.

\section{Experiments}
\label{sec:experiments}

\subsection{Dataset}
\begin{figure}[ht]
  \centering
  \includegraphics[width=0.92\linewidth]{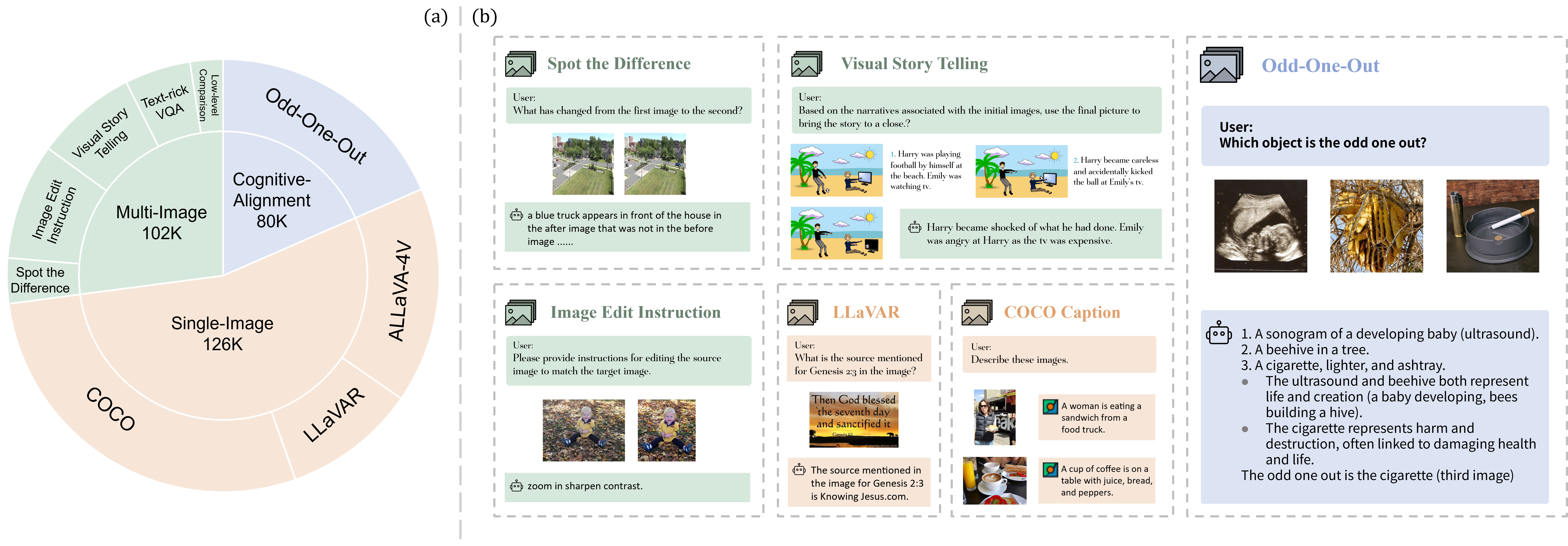}
  \caption{(a) Overview of hybrid dataset. The total dataset includes three subsets: Single-image (126K samples), Multi-image (102K samples), and Cognitive-Alignment (80K samples). (b) Examples. The examples include Spot the Difference, Image Edit Instruction, Visual Storytelling, LLaVAR, COCO Caption, and Odd-One-Out tasks.}
  \label{fig:hybrid-dataset-overview}
\end{figure}

We construct a hybrid dataset including $\mathcal{D}_{\text{O1O}}$ and $\mathcal{D}_{\text{gen}}$. The composition is summarized in Figure~\ref{fig:hybrid-dataset-overview} (a) and examples are shown in Figure~\ref{fig:hybrid-dataset-overview} (b). Details are provided in Appendix~\ref{apx:data-details}.

\subsection{Setup}
We conduct experiments across the Qwen3-VL \citep{qwen3vl} model family, specifically utilizing the $\sim$4B and $\sim$8B instruction models, as well as the $\sim$30B Mixture-of-Experts (MoE) model. We denote the aligned models fine-tuned via our framework as \textbf{CogAlign}, and their unaligned counterparts as \textbf{Baseline}. Full implementation details are provided in Appendix~\ref{apx:train-details}.

In addition to these MLLMs, we incorporate \textbf{Centaur}, a foundation model of human cognition. This serves as a specialized external reference, allowing us to benchmark our aligned models against an architecture optimized for cognitive tasks.

\subsection{Main Result}
\paragraph{Generalization performance.}
A critical concern in model alignment is the potential degradation of general capabilities. To investigate whether aligning with the latent structures of human cognition compromises broader reasoning abilities, we evaluate both the baseline and aligned models on two multimodal benchmarks: MMMU \citep{yue2024mmmu} (val) and MMBench \citep{liu2024mmbench} (en, test).

\begin{wraptable}{r}{0.54\linewidth}
    \centering
    \small
    \vspace{-13pt}
    \captionof{table}{Performance comparison of baseline and aligned models. CogAlign yields consistent accuracy improvements on $\text{MMMU}_\text{val}$, while incurring a slight performance degradation on $\text{MMBench}_{\text{en, test}}$ across all model scales. Details are provided in Table~\ref{tab:mmbench-details} (Appendix~\ref{apx:gen-performance-details}).}
    \label{tab:gen-performance}
    \setlength{\tabcolsep}{3pt}
    \renewcommand{\arraystretch}{1.2}
    \begin{tabular}{lcccc}
      \toprule
      \multicolumn{1}{c}{\multirow{1}{*}{Model}} & \multicolumn{2}{c}{MMMU$_{val}$ (\%)} & \multicolumn{2}{c}{MMBench$_{en,test}$ (\%)} \\
      Qwen3-VL
      & Baseline  & CogAlign  & Baseline & CogAlign\\
      \midrule
      4B & 48.44 & \textbf{51.56} & \textbf{82.96} & 81.39 \\
      8B & 50.67 & \textbf{54.22} & \textbf{83.91} & 83.86 \\
      30B-A3B & 53.22 & \textbf{57.56} & \textbf{87.44} & 83.86 \\
      \bottomrule
    \end{tabular}
    \vspace{-14pt}
\end{wraptable}

As reported in Table~\ref{tab:gen-performance}, we evaluate the general capabilities to investigate the potential alignment tax. The results indicate a trade-off: while the aligned models achieve consistent performance gains across all scales on $\text{MMMU}_{\text{val}}$ benchmark (e.g., +4.34 improvement for Qwen3-VL-30B), they exhibit a minor degradation on $\text{MMBench}_{\text{en, test}}$ compared to their unaligned baselines. This suggests that aligning with human cognitive latent structure incurs a slight alignment tax on general-purpose multimodal tasks.

\paragraph{Human consistency.}
We quantify human consistency using accuracy on the O1O validation set comprising 453,642 held-out triplets, where higher accuracy indicates better alignment. This massive scale ensures that our evaluation captures robust results rather than anecdotal success.

As detailed in Table~\ref{tab:human-consistency}, we evaluate the models against two distinct baseline settings to establish a comprehensive comparison. In the Direct prompt~\ref{prompt:o1o-direct} setting, the model is prompted to output the final option (e.g., A, B, or C) immediately. In the CoT prompt~\ref{prompt:o1o-cot-infer} setting, the model is required to think step-by-step before decision-making. The results demonstrate that our alignment framework yields improvements over both baseline settings and the performance of aligned models remains stable across all scales (ranging from $56.21$ to $56.99$). While the aligned models outperform not only the Direct baselines but also more capable CoT baselines across all models, the relative gain over the CoT baseline narrows as model capacity increases. This shrinking delta is due to the upward trajectory of the baseline itself.

\vspace{-0.24cm}
\begin{table}[h]
  \small
  \caption{Human consistency evaluation across baseline and aligned models. The baseline models are evaluated under two inference paradigms: Direct (predicting option immediately) and CoT (thinking step-by-step prior to decision-making). The proposed alignment framework consistently outperforms both the Direct and the more capable CoT baselines across all model scales (noise ceiling is 68.74 \citep{hebart2023things}).}
  \label{tab:human-consistency}
  \begin{center}
    \renewcommand{\arraystretch}{1.2}
    \begin{tabular}{lcccc}
      \toprule
      \multicolumn{2}{c}{\multirow{1}{*}{Model}} & \multicolumn{2}{c}{Baseline (\%)}  & \multirow{1}{*}{CogAlign (\%)} \\
      \midrule
      Centaur & 70B & \multicolumn{2}{c}{51.96} & --- \\
      \midrule
      \multicolumn{2}{c}{} & Direct & CoT & Ours \\
      \multirow{3}{*}{Qwen3-VL}
      & 4B & 41.48 $\pm$ 0.30 & 53.37 $\pm$ 0.30 & \textbf{56.81} $\pm$ 0.29 \\
      & 8B & 40.46 $\pm$ 0.29 & 52.28 $\pm$ 0.30 &\textbf{56.99} $\pm$ 0.29 \\
      & 30B-A3B & 37.09 $\pm$ 0.28 & 55.38 $\pm$ 0.30 & \textbf{56.21} $\pm$ 0.29 \\
      \midrule
      \multicolumn{2}{c}{Noise Ceiling} & \multicolumn{3}{c}{68.74 $\pm$ 1.07} \\
      \bottomrule
    \end{tabular}
  \end{center}
\end{table}

To rigorously validate these gains, we report 95\% Confidence Intervals (CIs) estimated via bootstrap resampling. For the Qwen3-VL series, the CIs of the aligned models (e.g., $[56.70, 57.28]$ for the $\sim$8B model) do not overlap with those of their respective baselines, indicating a statistically significant enhancement in human-model consistency by a McNemar test \citep{mcnemartest} ($p < 0.001$).

\paragraph{Neuroscientific analysis.}
To investigate whether our framework improves the alignment between the internal representational geometry of MLLMs and the human brain, we analyze the neural alignment shift ($\tilde{\Delta}_{RSA}(c)$). Following the method detailed in Section~\ref{method:neural-rsa}, this shift is computed using the concatenated hidden states from the MLLMs last four layers and visualized via cortical projection.

\begin{figure}[h]
  \centering
  \includegraphics[width=0.92\linewidth]{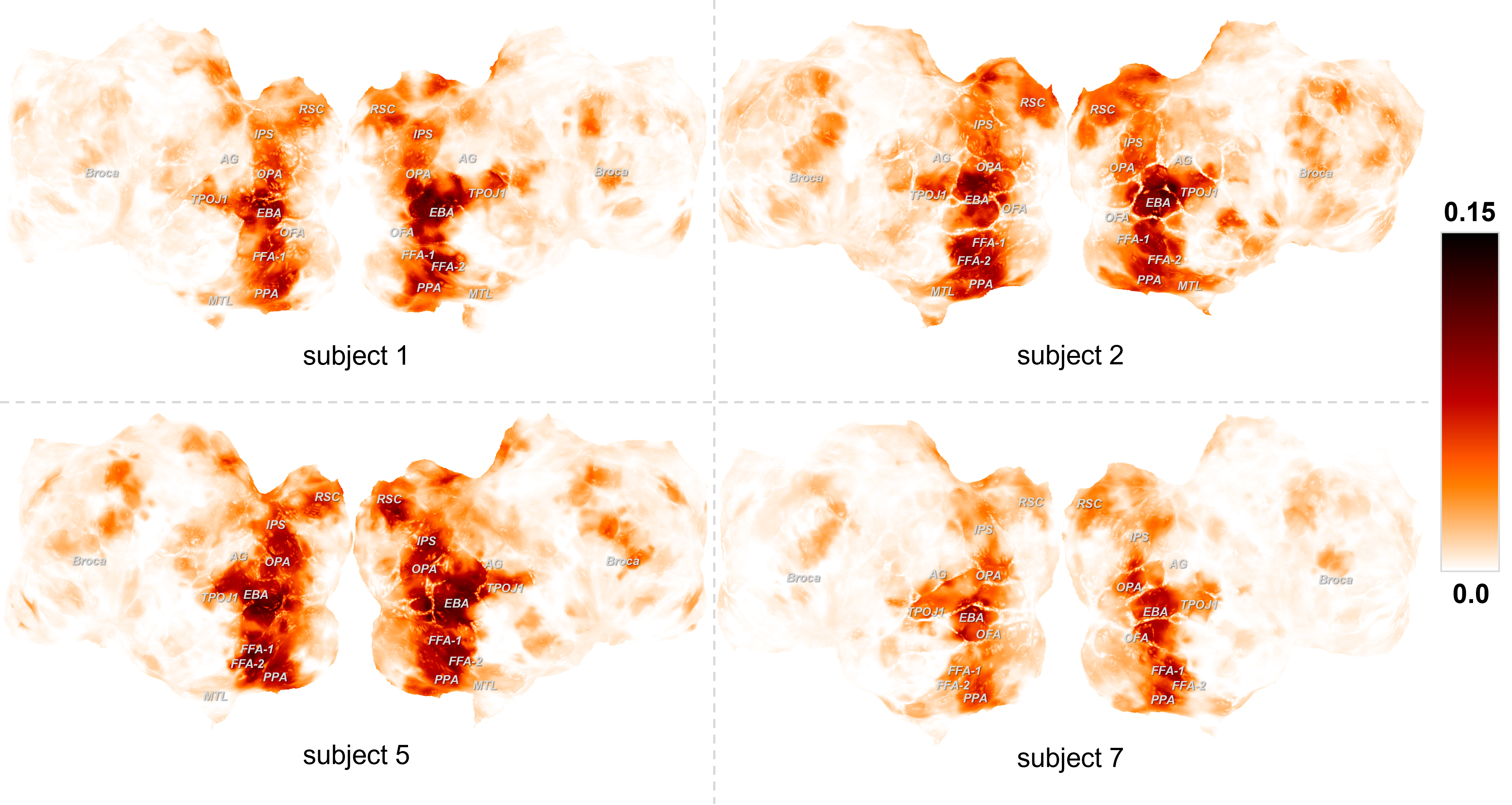}
  \caption{Cortical projection of representational alignment improvements across subjects. The heatmaps visualize the voxel-wise difference in $\tilde{\Delta}_{RSA}(c)$ scores between the aligned and baseline Qwen3-VL-4B-Instruct model. Darker colors indicate better representational alignment between models and human brain. Consistent positive shifts are observed, demonstrating that our alignment framework robustly shifts the MLLMs internal representational geometry to be more human-like.}
  \label{fig:subjs-4b-it-neuro-proj}
\end{figure}

\begin{figure}[h]
  \centering
  \includegraphics[width=0.92\linewidth]{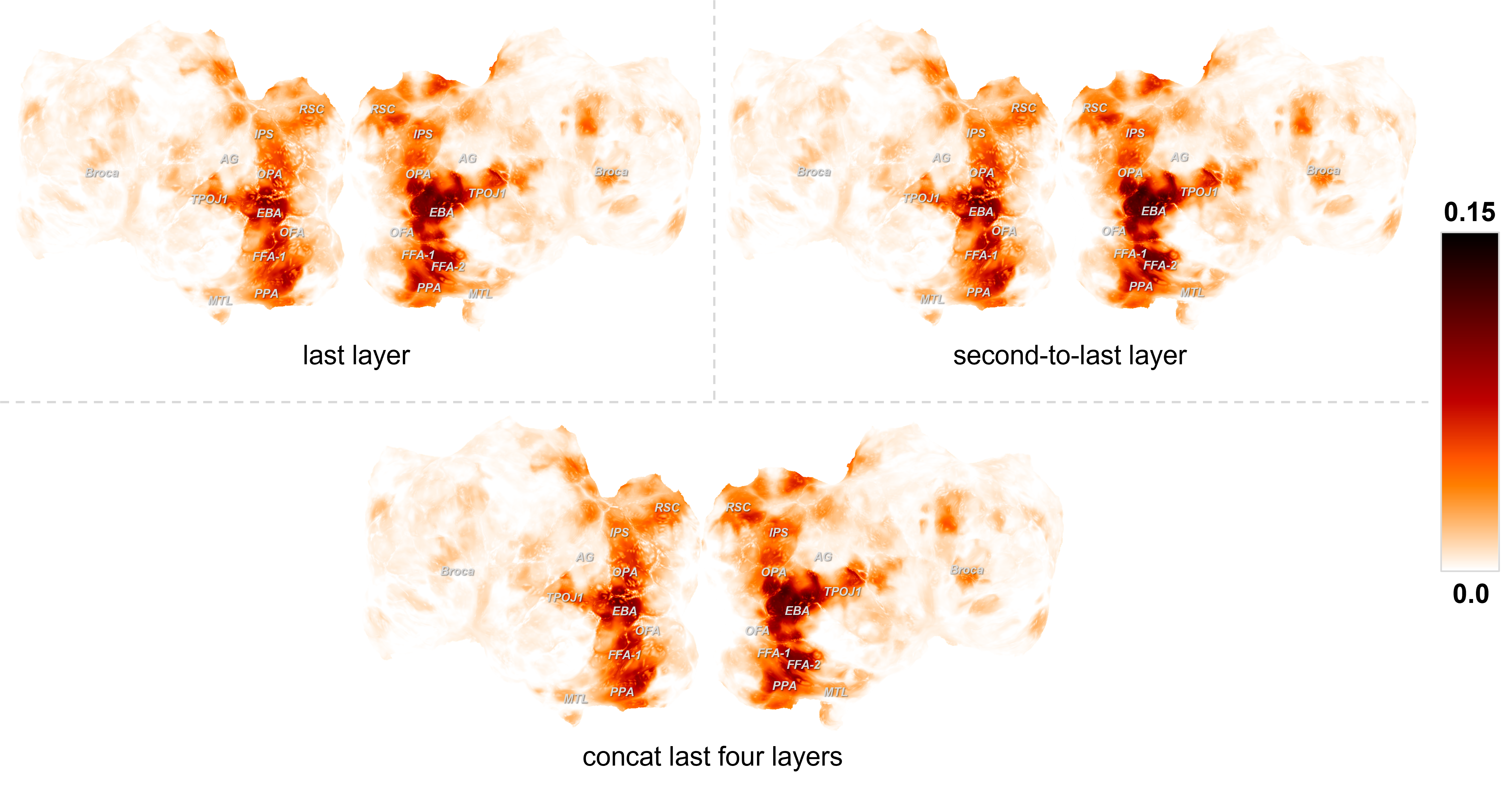}
  \caption{Stability of alignment shifts across different extraction depths. The cortical projections display the alignment improvements $\tilde{\Delta}_{RSA}(c)$ of subject 1 computed using features from the last layer, the second-to-last layer, and the concatenation of the last four layers. Darker regions reflect better alignment improvement. The spatial distribution of these positive shifts remains consistent regardless of the extraction depth, confirming that the achieved representational alignment with human cognition is robust and not a selective layer choice.}
  \label{fig:subj01-4b-it-hs-all-types}
\end{figure}

As illustrated in Figures~\ref{fig:subjs-4b-it-neuro-proj} and \ref{fig:subj01-4b-it-hs-all-types}, the resulting cortical heatmaps reveal consistent representational alignment improvements across all subjects (subject 1, 2, 5, and 7 \citep{yutakagi2023high,paulscotti2023reconstructing}; paired t-test, $p < 0.001$). Detailed discussion of these results is provided in the Appendix~\ref{apx:neural-details}.

\subsection{Ablation Result}
\paragraph{Necessity of alignment over prompting.}
To determine whether MLLMs can show human-like cognitive structures purely via inference-time prompting or if our alignment is necessary, we ablate the Qwen3-VL-4B-Instruct baseline against three prompting strategies: (1) Standard CoT (S-CoT~\ref{prompt:o1o-cot-infer}), which uses unconstrained step-by-step reasoning. (2) $d^*$-Guided CoT ($d^*$-CoT~\ref{prompt:o1o-cot-infer-dcot}), which strictly constrains reasoning to the most salient semantic dimension ($d^*$). (3) Full-Space CoT (FS-CoT~\ref{prompt:o1o-cot-infer-fscot}), which requires dynamic grounding one or more dimension(s) across the complete 66-dimensional behavioral semantic space.

\begin{wraptable}{r}{0.52\linewidth}
    \centering
    \small
    \vspace{-13pt}
    \captionof{table}{Prompting versus alignment. We compare three inference-time prompting strategies (S-CoT, $d^*$-CoT, and FS-CoT) with our alignment framework. Notably, providing semantic dimensions in the prompt during inference time negatively impacts model performance, whereas our CogAlign achieves the best overall results. The chance level is 33.33\%. See Appendix~\ref{apx:ablation-prompting} for details.}
    \label{tab:ablation-prompting}
    \setlength{\tabcolsep}{3pt}
    \renewcommand{\arraystretch}{1.2}
    \begin{tabular}{lccccc}
      \toprule
      \multicolumn{2}{c}{Model} & S-CoT & $d^*$-CoT & FS-CoT & CogAlign \\
      \midrule
      \multirow{1}{*}{Qwen3-VL}
      & 4B & 53.37 & 49.01 & 52.70 & \textbf{56.81} \\
      \bottomrule
    \end{tabular}
\end{wraptable}

As shown in Table~\ref{tab:ablation-prompting}, we compare inference-time prompting with our alignment framework on Qwen3-VL Instruct model. Although explicitly providing semantic dimensions enables the model to access these concepts during reasoning, it does not guarantee stable reasoning over these dimensions. Neither $d^*$-CoT nor FS-CoT consistently improves over standard reasoning (S-CoT). $d^*$-CoT reduces performance from 53.37 to 49.01, suggesting that imposing a cognitive dimension may constrain the model's decision process without ensuring effective utilization of the underlying semantic information. Similarly, FS-CoT achieves 52.70, remaining comparable to S-CoT. In contrast, our CogAlign achieves $56.81$, exceeding S-CoT and FS-CoT by $3.44$ and $4.11$ percentage points, respectively. Our framework incorporates these latent cognitive structures into the model, enabling more consistent human-like decision patterns.

\paragraph{Generalizability across MLLMs families.}
To verify that our alignment framework is architecture-agnostic, we extend our evaluation to Qwen2.5-VL \citep{qwen25vl} and Gemma3 \citep{gemma3} instruction models. To balance computational efficiency with reliability, we conduct this ablation on a subset ($\sim$75.6K triplets, 1/6) of the O1O validation set.

\begin{wraptable}{r}{0.52\linewidth}
    \centering
    \small
    \vspace{-13pt}
    \captionof{table}{Ablation on architectural generalizability.}
    \label{tab:human-consistency-mllms}
    \setlength{\tabcolsep}{3pt}
    \renewcommand{\arraystretch}{1.2}
    \begin{tabular}{lccc}
      \toprule
      \multicolumn{2}{c}{\multirow{1}{*}{Model}} & \multicolumn{1}{c}{Baseline (CoT)} & \multirow{1}{*}{CogAlign (Ours)} \\
      \midrule
      \multirow{1}{*}{Qwen2.5-VL} & 7B & 56.63 & \textbf{59.29} \\
      \multirow{1}{*}{Gemma3} & 12B & 55.77 & \textbf{57.28} \\
      \bottomrule
    \end{tabular}
    \vspace{-12pt}
\end{wraptable}

As reported in Table~\ref{tab:human-consistency-mllms}, our framework yields consistent improvements in human behavioral consistency across distinct model families. The aligned Qwen2.5-VL and Gemma3 outperform their respective CoT baselines by 2.66 and 1.51.

\paragraph{Similarity of behavioral representational geometry.}
Beyond agreement on triplet choices, we examine whether alignment captures the broader organization of human similarity judgements.

\begin{wrapfigure}{r}{0.36\linewidth}
    \centering
    \small
    \vspace{-13pt}
    \includegraphics[width=0.90\linewidth]{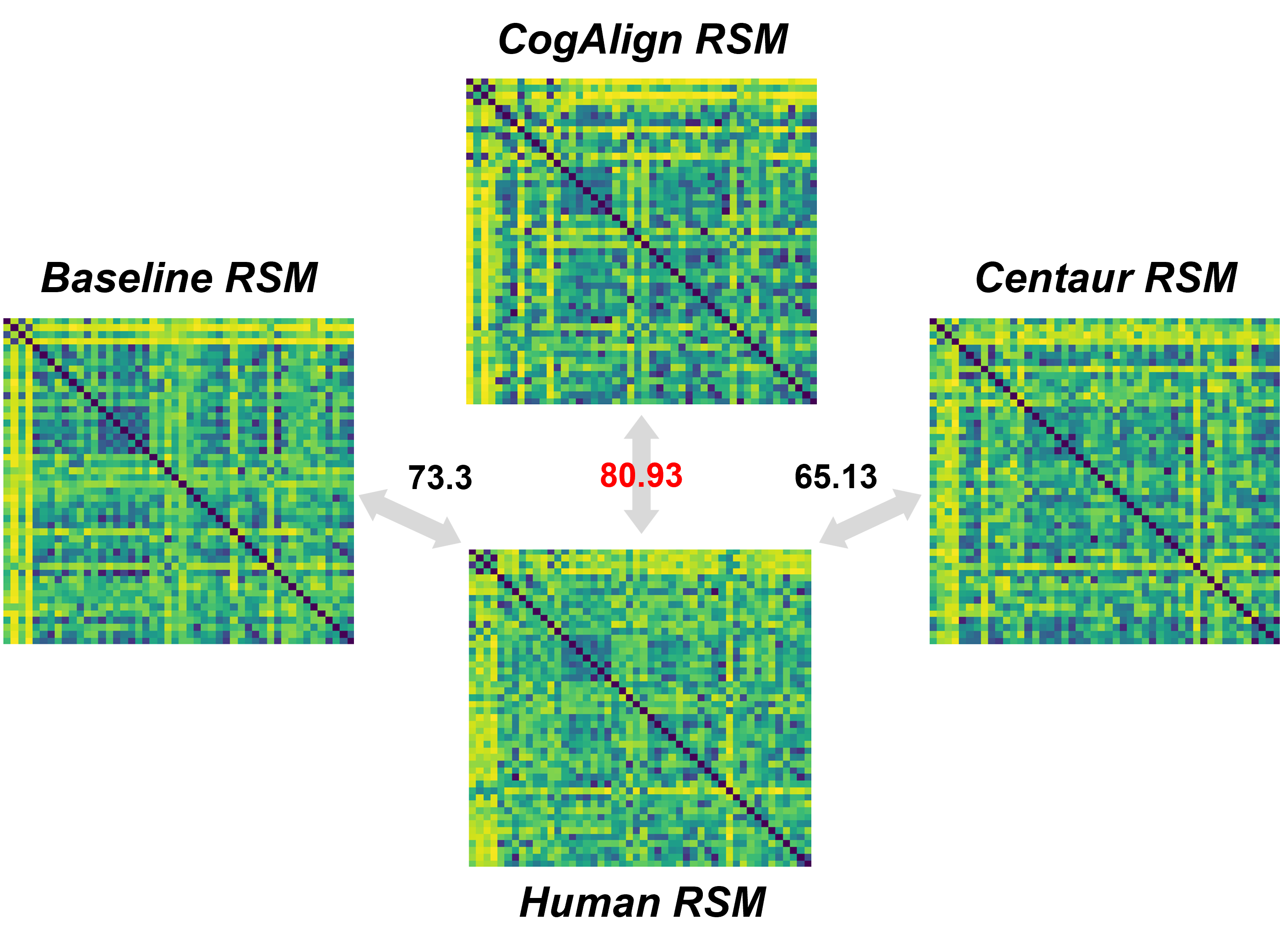}
    \captionsetup{hypcap=false}
    \captionof{figure}{Comparison of choice-based RSMs between Qwen2.5-VL-7B and human.}
    \label{fig:48objectrsa}
    \vspace{-10pt}
\end{wrapfigure}

Choice-based representational similarity matrices (RSMs) represent pairwise similarity relationships among objects, allowing us to assess model-human correspondence at the level of behavioral representational geometry. We compute the Pearson correlation of the choice-based RSMs derived from Qwen2.5-VL-7B and human on a 48-object subset ($\sim$91.56K triplets) drawn from the complete dataset. As shown in Figure~\ref{fig:48objectrsa}, the aligned model achieves 80.93\% exceeding its CoT baseline 73.30\% and Centaur 65.13\%. These results suggest that alignment improves correspondence with human behavioral representational geometry beyond triplet choices. Details are provided in Appendix~\ref{apx:48objectrsa}.
\section{Conclusion}
\label{sec:conclusion}

In this paper, we present a data-driven framework to align MLLMs with the latent structure of human cognition via behavior-derived semantic dimensions. Moving beyond outcome-level mimicry, our framework translates the multidimensional mental representations of natural objects underlying human similarity judgements into explicit linguistic rationales, enabling MLLMs to internalize these priors. Experimental evaluations demonstrate the efficacy of our alignment framework. The aligned MLLMs achieve improvements in behavioral consistency with human. Ablation studies reveal that these cognitive priors should be internalized: providing semantic dimensions via inference-time prompting paradoxically degrades performance, suggesting a fundamental gap between prompting and our framework. Furthermore, independent searchlight RSA indicates that the MLLMs internal representations become more aligned with human neural activity after aligning. While a minor trade-off is observed on general benchmarks, this work provides a proof-of-concept for bridging cognitive science and MLLMs alignment. Our framework points to a data-driven route for incorporating human cognitive structure into MLLMs alignment. We hope this work inspires further exploration toward developing more controllable, transparent, and human-like artificial cognitive systems. Furthermore, to clarify the commensurability of our baselines and explicitly outline the value of this process-level alignment framework for both the machine learning and cognitive science communities, we provide extended discussion in Appendix~\ref{apx:framework-value}.

\section{Limitations}
\label{sec:limitions}

While our framework improves alignment between MLLMs and the latent structure of human cognition, we note some limitations.

Our triplet-level semantic dimension attribution relies on a top-1 approximation, identifying a single dominant dimension ($d^*$). This isolates the primary semantic axis, but does not model the joint contributions of multiple dimensions or their context-dependent interactions. The resulting supervision may therefore simplify the structure underlying human similarity judgements.

The 66 behavior-derived semantic dimensions are object-centric, characterizing perceptual and conceptual properties primarily. Although this space includes some abstract dimensions, such as valuable or precious and weapon-related (Table~\ref{tab:66-dims}), their presence does not establish coverage of the abstractions required for mathematical reasoning or reasoning about complex geometric and topological structures. Consequently, the applicability of our alignment framework to highly abstract multimodal domains remains to be further explored (see Appendix~\ref{apx:abstract-dims-impact} for details).

As with many synthesized datasets, the rationales generated in our pipeline are not flawless. While our framework strictly grounds every rationale in a valid, human-derived cognitive dimension, the subsequent verbalization by the LLMs can occasionally introduce factual inconsistencies or ad hoc associations. We provide transparent and detailed discussion in Appendix~\ref{apx:rationale-quality}.

The improvement over the CoT baseline becomes smaller for larger models. This trend is driven by the emergent conceptual structures inherent in stronger models, which raises the initial untuned performance closer to our current alignment ceiling (see Appendix~\ref{apx:scaling-dynamics-ceiling} for details).

\begin{ack}
  This work was supported in part by Beijing Natural Science Foundation under Grant L243016; in part by the National Natural Science Foundation of China under Grant 62576336 and 82272072; and in part by Zhongguancun Academy (Grant No.s P210260301).
\end{ack}

{
  \small
  \bibliographystyle{unsrtnat}
  \bibliography{references}
}

\appendix
\section{Preliminaries}\label{sec:prel}

Prior experimental, neuropsychological and neuroimaging evidence have led to the proposal of object dimensions such as animacy, manipulability and real-world size \citep{alfonsocaramazza1998domainspecific,chao1999attributebased,konkle2011canonical}, but these dimensions describe only a selective and largely incomplete portion of our mental representational space. Our work builds on a previously established behavior-derived semantic embedding space, which provides visual and conceptual mental representations for a wide, representative range of natural objects \citep{hebart2020revealing,hebart2023things}. Specifically, this space maps a set of 1854 object concepts into a sparse and non-negative vector space, where each dimension corresponds to a distinct conceptual and perceptual property (e.g., metallic, animal-related, textile).

\subsection{Sparse Positive Similarity Embedding (SPoSE)}
We introduce the SPoSE proposed by Charles Y. Zheng et al. \citep{spose} in this section. SPoSE is based on three assumptions. 

\paragraph{Semantic representation of concepts.}
The first assumption is that each object concept $c_i$ can be represented by semantic features, quantified in a vector ${\bf x_i} = (x_{i1},\ldots,x_{ip})$. These features correspond dimensions of a  latent space where the geometrical relationships encode conceptual (dis)similarity. The SPoSE model estimates these features {\em from behavioral data alone} using the probabilistic model. To ensure interpretability, the SPoSE model constrains each dimension $x_{if}$ in the vector ${\bf x_i}$ to be real and non-negative, reflecting the degree to which the aspect of meaning it represents is present and influences subject behavior \citep{murphy2012learning}. Furthermore, a sparsity constraint is added to the SPoSE model for sparse these dimensions \citep{mcrae2005semanticfp}.

\paragraph{Probabilistic model.}
The second assumption is that the decision in a triplet task trial is explained by the similarity between the embedding vectors of the presented concepts. For any two concepts $c_i$ and $c_j$, the SPoSE model defines the similarity $S_{ij}$ as the dot product of their corresponding vectors:
\begin{equation}\label{eq:spose-similarity}
S_{ij} = \langle {\bf x_i}, {\bf x_j} \rangle = \sum_{f=1}^p x_{if} x_{jf}. 
\end{equation}

The third assumption models the outcome of the triplet task as a stochastic process. This assumption reflects both trial-to-trial inconsistencies in how an individual subject makes the decision, as well as subject differences in the decision-making process. To account for behavioral inconsistencies, the probability of a subject choosing a specific pair is modeled by Softmax Action Selection:
\begin{equation}
\Pr[y_1,y_2] \propto \exp(S_{y_1,y_2})
\end{equation}

Since probabilities of the three possible outcomes add to one, this gives

\begin{equation}\label{eq:triplet_probs}
\Pr[y_1,y_2] = \frac{\exp(S_{y_1,y_2})}{\exp(S_{i_1,i_2}) + \exp(S_{i_1,i_3}) + \exp(S_{i_2,i_3})}
\end{equation}

In this formulation, the vectors ${\bf x_i}$ are unnormalized. This design choice allows vector similarities to be arbitrarily large, modeling situations where decision is near-deterministic (e.g., when two concepts are virtually identical).

\paragraph{Parameter fitting.}
The embedding vectors $({\bf x_1},\hdots, {\bf x_m})$ are inferred by minimizing a regularized maximum likelihood objective function. For the given triplets datasets (total number is $n$), where $(i_{1,j}, i_{2,j}, i_{3,j})$ denotes the concepts in the $j$-th triplet and $(y_{1,j}, y_{2,j})$ indicates the chosen pair. The objective function to be minimized is defined as:

\begin{equation}\label{eq:sparse_triplet_objective}
\sum_{j=1}^{n_{train}} \log \left(\frac{\exp({\bf x_{y_{1,j}}}^T {\bf x_{y_{2,j}}})}{\exp({\bf x_{i_{1,j}}}^T {\bf x_{i_{2,j}}}) + \exp({\bf x_{i_{1,j}}}^T {\bf x_{i_{3,j}}}) + \exp({\bf x_{i_{2,j}}}^T {\bf x_{i_{3,j}}})}\right) + \lambda \sum_{i=1}^{m} ||{\bf x_i}||_1
\end{equation}

where $p$ is specified number of dimensions, $n_{train}$ is the number of training triplets. $||\cdot||_1$ is the L1 norm, so $||{\bf x}||_1 = \sum_{f=1}^p |x_f|$, and $x_f \ge 0$ for $f=1,\hdots,p$. $\lambda$ is the regularization parameter.

\subsection{Behavior-Derived Embedding Space via SPoSE}\label{preliminary:embedding}
The object representations are learned from human similarity judgement behavioral data using the SPoSE model. In the triplet O1O task, participants are presented with three objects, $i, j,$ and $k$, and asked to identify the most dissimilar one. Assuming $k$ is chosen as the odd-one-out, the psychological similarity between $i$ and $j$ is considered the highest among the pairs. SPoSE models this by assigning a non-negative latent vector $\mathbf{x} \in \mathbb{R}_{\ge 0}^d$ to each object. The probability of choosing $k$ as the odd-one-out is formulated using a softmax over the inner products of the object vectors: 
\begin{equation}
P(k \mid i,j,k) = \frac{\exp(\mathbf{x}_i^\top \mathbf{x}_j)}{\exp(\mathbf{x}_i^\top \mathbf{x}_j) + \exp(\mathbf{x}_i^\top \mathbf{x}_k) + \exp(\mathbf{x}_j^\top \mathbf{x}_k)}
\end{equation}

By applying $L_1$ regularization during optimization, SPoSE enforces sparsity and non-negativity, ensuring that the resulting dimensions are interpretable as the latent structure of human cognition \citep{hebart2020revealing}.

\subsection{Justification for 66 Dimensions}
The critical property of this embedding space is its dimensionality, $\textbf{d = 66}$. In many representation learning contexts, dimensionality is treated as an arbitrary hyperparameter tuned for downstream performance. However, in this behavior-derived space, $d=66$ is saturation. Analysis conducted by Hebart et al. \citep{hebart2023things} demonstrates that while the number of reproducible dimensions scales with the volume of behavioral data, it ultimately saturates around 67.5 dimensions and would reach $\sim 66.5$ dimensions for $4.5 - 5$ million trials. Indeed, when re-running the SPoSE model with the full dataset of 4.70 million trials (4.10 million for training), embedding dimensionality turns out as predicted: from a set of 72 randomly-initialized models, they choose the most reliable embedding as the final embedding, revealing 66 interpretable dimensions underlying perceived similarity judgements. Thus, increasing dataset size beyond this large dataset may no longer yield noticeable improvements in predictive performance or changes in embedding dimensionality at the global level of similarity, and potential improvements may not justify the cost of collecting additional data. These 66 dimensions represent a data-driven, globally saturated approximation of the latent structure of human cognition used to conceptualize this broad set of natural objects. This provides a rigorous and interpretable foundation for our proposed alignment framework.

\section{Impact of Incorporating Abstract Dimensions}
\label{apx:abstract-dims-impact}
Our alignment framework makes no assumptions about the semantic content of the dimensions. Incorporating richer abstract dimensions therefore requires replacing or augmenting the embedding space rather than modifying the core framework.

We envision three pathways for this integration: (a) Swapping the behavioral substrate. Applying the identical pipeline to O1O data collected on stimuli where abstract criteria dominate (e.g., social or moral scenarios). (b) Augmenting the existing space. Concatenating normative ratings for objects along abstract attributes with the SPoSE dimensions before Jackknife attribution, allowing the attribution step to decide whether an abstract or object-centric criterion best explains the human choice. (c) Context-conditioned judgements. Since abstract criteria often become decision-relevant only under specific contexts, collecting context-conditioned O1O data and injecting this context into the rationale prompt.

Abstract criteria (e.g., social acceptability) are substantially more subject- and culture-dependent, introducing variability into the decision-making process. While this could enhance the model's alignment with complex human values, it also raises the bias-auditing concerns we raise in the Impact Statement. As illustrated by the emergent stereotypical gender dimensions, behaviorally derived abstract axes can encode societal norms or biases that human may not wish to instill wholesale, necessitating careful curation when expanding the semantic space.

\section{Rationale Quality Analysis}
\label{apx:rationale-quality}
In our framework, the supervision signal originates from two distinct sources. (1) Human Population: The decisive supervisory anchor, the choice and its operative cognitive dimension, is attributed from 4.70M validated human judgements via SPoSE and Jackknife. (2) LLM Generator: The LLM role is strictly restricted to verbalizing this behaviorally grounded criterion without altering the underlying decision.

A concern regarding this pipeline is the quality of the LLM-generated rationales. Since no generation is flawless, we conduct a transparent diagnostic of the rationale quality, evaluate the robustness of the framework across different rationale generators, and clarify why the framework yields robust gains despite data imperfections.

\paragraph{Independent MLLM-as-judge diagnostic.}
To quantitatively evaluate rationale quality without self-preference bias, we employ a stronger, independent MLLM (Kimi-K3 \citep{kimik3}) as a rubric-based judge on a random sample of training rationales. Given the triplet images, the corresponding dimension, and the rationale, the judge scores four independent criteria:
\begin{itemize}
  \item \textbf{Dimension faithfulness (D):} The given dimension is the actual operative basis of the comparison.
  \item \textbf{Factual correctness (F):} Claims about the objects are visually supported or represent stable common knowledge.
  \item \textbf{Logical validity (L):} The stated premises consistently entail the odd-one-out choice.
  \item \textbf{Non-forcedness (N):} The dimension-object relation is conventional and discriminative, penalizing ad hoc associations.
\end{itemize}

As shown in Table~\ref{tab:rationale-quality}, for Qwen2.5-72B-Instruct generator, the independent MLLM-as-judge diagnostic reveals that while 93.37\% of rationales are faithfully grounded in the assigned dimension, only about 70.45\% are strictly logically valid, 73.11\% are factually flawless, and 54.73\% are completely non-forced. Despite these verbalization imperfections, our results confirm that the framework consistently improves alignment across multiple MLLMs and generator LLMs.

\begin{table}[ht]
  \caption{Rationale quality across generators.}
  \label{tab:rationale-quality}
  \begin{center}
    \renewcommand{\arraystretch}{1.2}
    \begin{tabular}{lcccc}
      \toprule
      \textbf{Generator} & D & F & L & N \\
      \midrule
      Qwen2.5-72B-Instruct (Ours)  & 93.37 & 73.11 & 70.45 & 54.73 \\
      DeepSeek-V4-Flash            & 96.21 & 81.06 & 69.89 & 62.12\\
      \bottomrule
    \end{tabular}
  \end{center}
\end{table}

\paragraph{Cross-generator semantic convergence and robustness.}
To ensure that the framework effectiveness does not rely on the idiosyncrasies of a single generator (Qwen2.5-72B \citep{qwen25}), we regenerate the full set rationales using DeepSeek-V4-Flash \citep{deepseekv4} under identical prompts.

\textbf{Semantic convergence:} Using Qwen3-Embedding-0.6B \citep{qwen3embedding}, we compute the pairwise semantic similarity between rationales produced by the two generators for identical triplets. The high similarity (Mean = 83.39\%, Std = 5.93\%) confirms that different LLMs converge on similar reasoning pathways when anchored by the same cognitive dimension.

\textbf{Performance robustness:} We fine-tune the Qwen3-VL-4B-Instruct using the generated DeepSeek-V4-Flash rationales. The accuracy (56.52\%) closely matches our original result using Qwen2.5-72B rationales (56.81\%), and exceeds the CoT baseline (53.37\%).

\paragraph{Why imperfect rationales drive consistent improvement.}
A question arises: if a substantial proportion of training rationales are identified as flawed or forced, why should they be included, and why do they still drive performance improvements?

We argue that this highlights a core strength of our approach. The foundational anchor of every sample remains a valid human consensus. Even when the LLM verbalization is imperfect or slightly forced, it still forces the model to map its internal representations toward a validated human cognitive dimension rather than relying on uninterpretable statistical shortcuts. Crucially, our experiments demonstrate that the performance gains are the property of this dual-supervision framework. The exact same unfiltered dataset yields consistent improvements across five different MLLMs configurations spanning multiple scales and architectures (e.g., Qwen3-VL, Qwen2.5-VL, Gemma3 in Tables \ref{tab:human-consistency} and \ref{tab:human-consistency-mllms}). The fact that alignment gains replicate across multiple architectures and independent rationale generators, despite the quantified imperfections, provides strong evidence that the improvement stems from the structure of the framework itself rather than data curation.

While employing a closed-loop validation procedure or a frontier model (like Kimi-K3) to regenerate and filter the dataset would undoubtedly improve data quality and aligned model performance, such large-scale regeneration is often economically prohibitive for exploratory studies. Our findings confirm that even with accessible open-weight generators and inherently noisy verbalizations, injecting human cognitive priors explicitly via this framework improves alignment.

\section{Scaling Dynamics and Alignment Ceiling Analysis}
\label{apx:scaling-dynamics-ceiling}
As observed in our main results (Table~\ref{tab:human-consistency}), the relative accuracy gain of our aligned model over the CoT baseline diminishes as the parameter size of the MLLM increases. We decompose this scaling trend to provide more detailed discussion.

Across all model scales, the absolute performance of our aligned models remains remarkably stable: 56.81 (4B), 56.99 (8B), and 56.21 (30B). The shrinking delta is driven by the rising CoT baseline. Stronger MLLMs emerge human-like conceptual structures and world knowledge during large-scale pre-training. Consequently, the untuned starting point of larger models naturally moves closer to the human-like behavioral target, leaving less accessible headroom for our current alignment formulation. Currently, our pipeline only relies on top-1 dimension, supervising the model with a single dominant criterion. However, human similarity judgements are often nuanced, involving complex trade-offs among multiple cognitive dimensions. Supervising only the primary axis creates an implicit upper bound on task performance. Furthermore, our current alignment dataset comprises 80K sampled triplets, representing only  $\sim$2\% of the 4.1M available training triplets.

This yields a concrete, testable prediction for future alignment research: enriching the supervision signal through multidimensional rationales and scaling the alignment data far beyond the current $\sim$2\% sample should effectively break the ceiling. As model capabilities increase, they possess the representational capacity to absorb these denser, multi-faceted cognitive priors.

\section{Framework Scope and Interdisciplinary Value}
\label{apx:framework-value}
We further clarify the conceptual scope of our alignment framework, delineate its distinction from existing baselines, and outline its specific utility for researchers.

\paragraph{Baseline commensurability and process-level internalization.}
A core contribution of our work is shifting the alignment pathway for MLLMs from outcome-level behavioral mimicry to the process-level internalization of latent human cognitive structures. This distinction determines which baselines are commensurable with our approach. Post-hoc feature reweighting and selection methods operate as readouts on frozen feature spaces, producing task-specific discriminative mappings. They do not modify the model's internal reasoning and they are difficult to apply to the autoregressive generative setting of MLLMs. Consequently, the appropriate same-class baseline is a behaviorally fine-tuned model.

As reported in our main results, we compare against Centaur, which is fine-tuned on the largest existing human behavioral corpus (Psych-101). Our aligned models outperform Centaur both behaviorally and in representational similarity in Table~\ref{tab:human-consistency} and Figure~\ref{fig:48objectrsa}. Furthermore our prompting ablations in Table~\ref{tab:ablation-prompting} function as a non-parametric analogue to feature selection. This demonstrates that cognitive criteria should be internalized through joint rationale-decision supervision, rather than merely provided as external features.

\paragraph{Broad use cases and value for researchers.}
Beyond empirical performance, our framework offers a general, data-efficient recipe for injecting population-level cognitive priors into MLLMs without requiring architectural modifications.

For LLMs/MLLMs researchers, the primary value lies in intrinsic interpretability and controllability. Because every decision made by the model is preceded by a rationale grounded in an interpretable, behaviorally validated dimension, the decision processes become directly auditable against known human criteria. Moreover, the proposed framework is domain-agnostic. It enables researchers to convert existing large-scale cognitive science datasets into rich process-level supervision, facilitating the development of controllable, consensus-consistent LLMs/MLLMs.

For neuroscientists and cognitive scientists, the models additionally offer a new class of testable computational models and can be subjects for hypothesis generation about related tasks to reduce the difficulty and cost of data sampling. Our work also encourages relevant researchers to shift their focus from outcome-level studies to a deeper exploration of the underlying cognitive reasoning processes.

\section{Method Details}

\subsection{Internal Feature Extraction via Forward Propagation}
\label{apx:feature-extraction-method}
To map the MLLMs internal representations to human brain activity, we construct a multimodal input using a neutral instruction prompt (\texttt{Describe this image}) alongside the visual stimuli and perform forward propagation pass through baseline and aligned MLLMs. Targeting the most representative feature, we extract the hidden states specifically from the \texttt{<|vision\_end|>} token which is the special token delimiting the end of the visual embedding sequence for Qwen3-VL series models. We extract these states from three specific depths: the last layer, the second-to-last layer, and the last four layers \citep{bert}. We prioritize this token over individual spatial image pad tokens or subsequent text tokens for two fundamental reasons: (1) Global Visual Integration: Unlike individual image pad tokens which encode spatially localized features, the \texttt{<|vision\_end|>} token serves as a global aggregator. Due to the causal attention mechanism of the LLM backbone, this token attends to the entire preceding visual sequence, effectively summarizing the holistic visual semantic content before text begins. (2) Vision-Language Boundary: This token represents the \textit{terminal state} of visual perception and the \textit{initial state} of text. It is the bottleneck where visual features are projected into the semantic manifold, making it the most theoretically plausible candidate for alignment with human cortical representations.

\section{Data Details}
\label{apx:data-details}

\subsection{Statistics}
\label{apx:data-statistics}

\begin{table}[ht]
  \caption{Hybrid dataset composition.}
  \label{tab:data-statistics}
  \centering
  \renewcommand{\arraystretch}{1.4}
  \begin{tabular}{llcc}
    \toprule
    \textbf{Task} & \textbf{Dataset} & \textbf{Samples} & \textbf{Ratio} (\%) \\
    \midrule
    \multicolumn{4}{l}{\textbf{Single-image Scenarios}} \\
    Caption (13.5K)     & COCO           & 13,501 & 4.36 \\
    ALLaVA-4V (69.98K)  & ALLaVA-VFLAN   & 19,990 & 6.46 \\
    & ALLaVA-LAION   & 49,990 & 16.16 \\
    LLaVAR (43.17K)    & LLaVAR-GPT4    & 43,167 & 13.96 \\
    \addlinespace
    \midrule
    \multicolumn{4}{l}{\textbf{Multi-image Scenarios}} \\
    Spot the Difference (14.66K) & Spot-the-Diff & 10,774  & 3.48 \\
    & CLEVR-Change    & 3,885  & 1.26   \\
    Image Edit Instruction (23.013K) & MagicBrush   & 19,557 & 6.32 \\
    & IEdit        & 3,456  & 1.12   \\
    Visual Storytelling (32.94K)   & AESOP        & 6,915  & 2.24   \\
    & VIST        & 26,026  & 8.41   \\
    Text-rich VQA (21.39K)          & WebQA        & 9,338  & 3.02   \\
    & TQA          & 8,249  & 2.67   \\
    & OCR-VQA      & 1,899  & 0.61   \\
    & DocVQA       & 1,900  & 0.61    \\
    Low-level Comparison (10.68K)   & DreamSim     & 10,682  & 3.45 \\
    \addlinespace
    \midrule
    \multicolumn{4}{l}{\textbf{Odd-One-Out Task}} \\
    Cognitive-Alignment (80K)                        & THINGS       & 80,000 & 25.86 \\
    \midrule
    Total &  & 309329 & 100.0 \\
    \bottomrule
  \end{tabular}
\end{table}

Table~\ref{tab:data-statistics} summarizes the composition of our hybrid training dataset, reporting sample counts and proportions for cognitive-alignment data and generalization data across single-image and multi-image tasks.

\subsection{Cognitive-Alignment Data}
O1O task data to align the latent structure of human cognition. $\mathcal{D}_{\text{O1O}}$ is constructed via method~\ref{method:rationale-generation} using prompt~\ref{prompt:o1o-qa}.

\subsection{Generalization Data}
We supplement $\mathcal{D}_{\text{O1O}}$ with a diverse set of regularization data $\mathcal{D}_{\text{gen}}$ selected from open-source datasets, categorized into Single-image and multi-image scenarios:

\paragraph{Single-image scenarios.}
We incorporate COCO \citep{coco}, ALLaVA-4V \citep{chen2024allava}, and LLaVAR \citep{zhang2023llavar}.

\paragraph{Multi-image scenarios.}
Since the Odd-One-Out task inherently involves comparing multiple images (a triplet), we specifically include multi-image reasoning tasks from M4-Instruct \citep{li2024llava}, such as Spot the Difference \citep{jhamtani2018learning,park2019clevrchange}, Image Edit Instruction \citep{zhang2023magicbrush,bodur2024iedit}, Visual Storytelling \citep{ravi2021aesop,huang2016visual}, Text-rich VQA, and Low-level Comparison \citep{dreamsim,dreamsim2}. This ensures that the model learns to process multi-image text sequences robustly, rather than overfitting to the specific three-image format of the O1O task.

\subsection{Dynamic Stratified Data Sampling}
\label{apx:data-sampling-strategy}
Given the diverse nature of our training mixture (ranging from O1O cognitive data to general data), utilizing a naive data loader could lead to distributional shifts between batches (e.g., a batch dominated solely by one task type). To mitigate this, we implement a dynamic stratified sampling strategy.

Specifically, we define the sampling probability for each task $k$ based on its proportion in the total dataset. During training, for every global step, we construct the mini-batch by sampling data points from each task category proportional to these ratios, constrained by the total global batch size across all GPUs. The samples within each constructed batch are then shuffled using a fixed random seed.

This strategy guarantees that every gradient update vector is computed from a representative distribution of both cognitive alignment signals and general regularization data. By stabilizing the gradient direction and preventing task-specific bias during optimization, this sampling method is helpful for retaining the model's generalization capabilities while effectively instilling the latent structure of human cognition.

\section{Experimental Details}
\label{apx:experimental-details}
We provide a comprehensive overview of our experimental setup, including the hyperparameters for Supervised Fine-Tuning (SFT).

\subsection{Training Details}
\label{apx:train-details}

\paragraph{Supervised fine-tuning (SFT).}
As shown in Table~\ref{tab:sft-params}, the SFT stage utilizes AdamW \citep{loshchilov2019decoupled} and a cosine learning rate scheduler with a peak learning rate of $1\text{e-}5$. We total use 6 $\times$ NVIDIA A800 80G GPUs with TRL \citep{trl}, Qwen-VL utils \citep{qwenvlgithub}, and DeepSpeed \citep{zero}.

\begin{table}[ht]
  \centering
  \begin{minipage}[t]{0.32\textwidth}
    \centering
    \caption{SFT params.}
    \label{tab:sft-params}
    \resizebox{\linewidth}{!}{%
      \begin{tabular}{lc}
        \toprule
        \textbf{Parameter} & \textbf{Value} \\
        \midrule
        Optimizer    & AdamW   \\
        Learning Rate& 1e-5    \\
        Scheduler    & cosine  \\
        Weight Decay & 0.0     \\
        Epochs       & 1       \\
        Warmup Ratio & 3\%     \\
        Batch Size   & 144     \\
        Seed         & 42      \\
        \bottomrule
      \end{tabular}%
    }
  \end{minipage}
  \hspace{0.8cm}
  \begin{minipage}[t]{0.32\textwidth}
    \centering
    \caption{Eval params.}
    \label{tab:eval-params}
    \resizebox{\linewidth}{!}{
      \begin{tabular}{lc}
        \toprule
        \textbf{Parameter} & \textbf{Value} \\
        \midrule
        Engine       & vLLM     \\
        Precision    & BF16     \\
        Temperature  & 0.0      \\
        Top-p        & 1.0      \\
        Top-k        & -1       \\
        Rep. Penalty & 1.05     \\
        Max Tokens   & 8192     \\
        \bottomrule
      \end{tabular}
    }
  \end{minipage}
\end{table}

\subsection{Evaluation Details}
We conduct a comprehensive evaluation using the vLLM~\citep{vllm}. For evaluation metrics, we adopt traditional exact string matching to accurately assess response correctness about human consistency. Following the guidelines from the official vLLM documentation, we set the temperature to $0.0$, top-p to $1.0$, and top-k to $-1$. We apply a repetition penalty of $1.05$ and limit the maximum response length to $8192$ tokens.

To ensure a strictly fair comparison, we re-evaluate all baselines on our local infrastructure rather than relying on reported metrics. We observe a performance discrepancy between local reproduction and reported results, a known issue in benchmark evaluation~\citep{ardalanm2025vllmreproissue}. Therefore, we report results for both the Baseline and our CogAlign models under the exact same experimental setting. Specifically, we employ the lmms-eval framework~\citep{zhang2025lmms,boli2024lmmseval} and adhere strictly to its default parameters for MMMU and MMBench benchmarks. By using the standard default parameters, we eliminate potential biases arising from hyperparameter tuning.

\subsection{Generalization Performance Details}
\label{apx:gen-performance-details}

\begin{table}[ht]
  \centering
  \renewcommand{\arraystretch}{1.2}
  \caption{Detailed performance comparison of MMBench (en,test).}
  \label{tab:mmbench-details}
  \begin{tabular}{lcccccc}
    \toprule
    & \multicolumn{6}{c}{Qwen3-VL} \\
    \multirow{3}{*}[14pt]{Model} & \multicolumn{2}{c}{4B} & \multicolumn{2}{c}{8B} & \multicolumn{2}{c}{30B-A3B} \\
    & Baseline & CogAlign & Baseline & CogAlign & Baseline & CogAlign \\
    \toprule
    \multicolumn{1}{l}{\textit{\textbf{A. Overall}}} & 82.96 & 81.39 & 83.91 & 83.86 & 87.44 & 83.86 \\
    \multicolumn{7}{l}{\textit{\textbf{B. General Capabilities}}} \\
    AR & 82.99 & 85.76 & 80.90 & 87.85 & 87.15 & 87.50 \\
    CP & 87.58 & 85.44 & 89.08 & 87.15 & 88.65 & 86.30 \\
    FP-C & 82.19 & 74.09 & 78.95 & 74.90 & 85.43 & 81.78 \\
    FP-S & 87.94 & 87.69 & 90.20 & 90.20 & 93.72 & 88.69 \\
    LR & 71.68 & 67.63 & 73.99 & 74.57 & 79.77 & 73.41 \\
    RR & 73.46 & 74.41 & 78.67 & 77.25 & 81.99 & 75.36 \\
    \multicolumn{7}{l}{\textit{\textbf{C. Specific Tasks}}} \\
    Action Recognition & 93.18 & 90.91 & 92.05 & 89.77 & 93.18 & 93.18 \\
    Attribute Comparison & 85.90 & 80.77 & 87.18 & 80.77 & 88.46 & 87.18 \\
    Attribute Recognition & 90.91 & 91.92 & 90.91 & 93.94 & 94.95 & 90.91 \\
    Celebrity Recognition & 85.59 & 87.29 & 89.83 & 88.98 & 93.22 & 88.14 \\
    Function Reasoning & 87.74 & 87.74 & 87.74 & 90.57 & 91.51 & 90.57 \\
    Future Prediction & 75.00 & 73.61 & 76.39 & 77.78 & 80.56 & 73.61 \\
    Identity Reasoning & 96.34 & 96.34 & 96.34 & 97.56 & 98.78 & 97.56 \\
    Image Emotion & 77.11 & 73.49 & 74.70 & 72.29 & 80.72 & 75.90 \\
    Image Quality & 85.88 & 82.35 & 90.59 & 88.24 & 88.24 & 88.24 \\
    Image Scene & 92.25 & 91.47 & 95.35 & 93.80 & 91.47 & 89.92 \\
    Image Style & 83.53 & 80.00 & 84.71 & 80.00 & 83.53 & 77.65 \\
    Image Topic & 96.47 & 96.47 & 96.47 & 97.65 & 97.65 & 97.65 \\
    Nature Relation & 77.78 & 80.25 & 85.19 & 82.72 & 90.12 & 75.31 \\
    Object Localization & 81.73 & 78.85 & 84.62 & 83.65 & 90.38 & 81.73 \\
    OCR & 96.10 & 94.81 & 97.40 & 96.10 & 97.40 & 96.10 \\
    Physical Prop. Reas. & 67.00 & 75.00 & 61.00 & 77.00 & 73.00 & 76.00 \\
    Physical Relation & 51.92 & 57.69 & 48.08 & 51.92 & 57.69 & 65.38 \\
    Social Relation & 83.33 & 79.49 & 92.31 & 88.46 & 89.74 & 82.05 \\
    Spatial Relationship & 66.67 & 49.38 & 56.79 & 53.09 & 74.07 & 64.20 \\
    Struct. Image-Text & 69.31 & 63.37 & 72.28 & 72.28 & 79.21 & 73.27 \\
    \bottomrule
  \end{tabular}
\end{table}

Table~\ref{tab:mmbench-details} reports detailed results on the MMBench(en,test) for the baseline and CogAlign variants of Qwen3-VL-4B, Qwen3-VL-8B, and Qwen3-VL-30B-A3B. In addition to overall scores, it provides breakdowns by general capability and specific task, allowing a finer-grained comparison of performance before and after cognitive alignment.

\newpage

\subsection{Necessity of Alignment over Prompting}
\label{apx:ablation-prompting}
We use a more lenient parser for $d^*$-CoT and FS-CoT settings because of their generation failures and manually examine the parsing failures. We find that almost all failures are not caused by minor formatting variations or an overly strict parser. Instead, the model frequently enters a highly repetitive generation loop, repeatedly enumerating irrelevant positive and negative attributes. Consequently, it exhausts the maximum generation length before producing the required closing tag. Thus, such cases represent genuine generation failures rather than correct predictions that are rejected solely because of formatting. We list an example of parsing failures below.
\begin{tcolorbox}[
    breakable,
    colframe=gray!75, colback=black!2,
    width=\textwidth
  ]
  \small
  <thinking>

  First, let's analyze each object using the provided semantic dimensions.

  Object A: A rabbit. This is an animal. It is soft, fluffy, and related to nature. It is not metallic, not artificial, not food-related, not human-related, not tool-related, not plant-related, not house-related, not transportation-related, not circular, not flat, not long or thin, not pointed, not flying, not colorful, not white, not green, not yellow, not orange, not ....
\end{tcolorbox}

\subsection{Necessity of Dimension-Grounded Rationales}
To isolate the exact drivers of our framework alignment gains, we conduct ablations on the content of the supervision signal. First, replacing the rationale with a sparse, single-word dimension label severely degrades performance as shown in Table~\ref{tab:ablation-word-guide}. Second, fine-tuning on choice-only and generic CoT rationales fails to match the performance of our dimension-grounded rationales in Table~\ref{tab:ablation-comparison-supervisions}, confirming that the gains stem specifically from the injection of human cognitive priors rather than task-specific fine-tuning.

\begin{table}[ht]
  \caption{Ablation on supervision format during fine-tuning Qwen3-VL-4B-Instruct.}
  \label{tab:ablation-word-guide}
  \begin{center}
    \renewcommand{\arraystretch}{1.2}
    \begin{tabular}{lccc}
      \toprule
      Metric & Baseline (CoT) & Dimension-only & CogAlign (Ours) \\
      \midrule
      Accuracy & 53.37 & 42.81 & \textbf{56.81} \\
      Choice-based RSA & 81.79 & 44.01 & \textbf{85.01} \\
      \bottomrule
    \end{tabular}
  \end{center}
\end{table}

\begin{table}[ht]
  \caption{Comparison of Choice-only, Generic CoT, and ours supervision signals.}
  \label{tab:ablation-comparison-supervisions}
  \begin{center}
    \renewcommand{\arraystretch}{1.2}
    \begin{tabular}{lcccc}
      \toprule
      Model & Baseline (CoT) & Generic CoT & Choice-only & CogAlign (ours) \\
      \midrule
      Qwen2.5-VL-7B-Instruct & 56.63 & --- & 56.21 & \textbf{59.29} \\
      Qwen3-VL-4B-Instruct & 53.37 & 47.57 & --- & \textbf{56.81} \\
      \bottomrule
    \end{tabular}
  \end{center}
\end{table}

\subsection{Neuroscientific Analysis Details}
\label{apx:neural-details}
Positive RSA shifts are observed across the ventral occipito-temporal cortex, encompassing category-selective regions such as the EBA, FFA, and PPA. This indicates that the aligned MLLMs representational geometry better captures the high-level semantic organization for object and category discrimination. Because these alignment gains are localized to high-level visual cortices, they reflect an abstraction of category- and shape-relevant semantic structures, rather than mere low-level retinotopic or pixel-wise feature matching.

Furthermore, we observe substantial alignment shifts extending into the IPS and RSC. Given the triplet comparison nature of the O1O task, the engagement of the IPS aligns with its established role in the dorsal attention and multiple-demand networks, facilitating top-down attentional selection and relational comparison. Concurrently, shifts within the RSC likely indicate enhanced alignment with memory-linked contextual and scene-associative representations. Taken together, these neural effects demonstrate that our framework drives the MLLMs to capture behaviorally relevant, higher-order similarity structures. Beyond behavioral task performance, these results provide a convergent, neural-level validation: anchoring MLLMs to behavior-derived semantic dimensions explicitly steers their internal representations toward human-like cognitive geometry.

\subsection{Similarity of Behavioral Representational Geometry Details}
\label{apx:48objectrsa}
To assess whether the global decision-making structure of the models aligns with human cognition, we perform RSA based on the behavioral choices in the O1O task. This section details the construction of the choice-based RSMs and the subsequent calculation of the geometry similarity scores.

\paragraph{Choice-based RSM.}
We evaluate the representational geometry on a subset of $N=48$ objects, yielding an $N \times N$ matrix for human and evaluating models (Baseline, CogAlign, and Centaur). We initialize an all-zero similarity matrix $S \in \mathbb{R}^{N \times N}$.

The evaluation dataset consists of approximately 91.56K triplets drawn from this 48-object subset. For each triplet iteration $t = \{o_i, o_j, o_k\}$, the models are tasked with selecting the odd-one-out object using CoT. Let the chosen odd-one-out be $o_k \in t$. The remaining two objects, denoted as $(o_i, o_j)$, are deemed the most similar pair within the context of this triplet. We then update the similarity matrix by incrementing the co-occurrence counts for this pair $(i,j)$ and $(j,i)$:

$$S_{i,j} \leftarrow S_{i,j} + 1 \quad \text{and} \quad S_{j,i} \leftarrow S_{j,i} + 1$$

This process is repeated across all triplets in the subset. The resulting matrix $S$ captures the global behavioral representational geometry of the models. Using this method, we derive the human RSM ($S_{\text{human}}$) and model RSM ($S_{\text{model}}$).

\paragraph{Quantifying representational geometry similarity.}
We extract the upper triangular portion of their respective symmetric matrices, excluding the main diagonal. Let $\mathbf{v}_{\text{human}} \in \mathbb{R}^{D}$ and $\mathbf{v}_{\text{model}} \in \mathbb{R}^{D}$ be the flattened 1D vectors derived from the upper triangles of $S_{\text{human}}$ and $S_{\text{model}}$ respectively, where $D = N(N-1)/2$. The behavioral representational geometry similarity is quantified using the Pearson correlation:

\begin{equation}
  \text{Similarity Score} = \rho(\mathbf{v}_{\text{model}}, \mathbf{v}_{\text{human}})
\end{equation}

where $\rho(\cdot, \cdot)$ is the Pearson correlation coefficient across the $D$ feature dimensions.


\section{Behavior-Derived 66 Semantic Dimensions}
As shown in Table~\ref{tab:66-dims}, we provide the saturated mental representations of natural objects underlying human similarity judgements. They are represented by \textbf{66} semantic dimension concepts.

\begin{table}[!htbp]
  \centering
  \caption{Overview of 66 semantic dimension concepts.}
  \label{tab:66-dims}
  \small
  \renewcommand{\arraystretch}{1.4}
  \setlength{\tabcolsep}{5pt}
  \begin{tabular}{lclc}
    \toprule
    \textbf{Axis} &\textbf{Semantic Concept} &\textbf{Axis} &\textbf{Semantic Concept} \\
    \midrule
    0 & metallic or artificial & 1 & food-related \\
    2 & animal-related & 3 & textile \\
    4 & plant-related & 5 & house-related or furnishing-related \\
    6 & valuable or precious & 7 &transportation-related or movement-related \\
    8 &body-related or people-related & 9 &wood-related or brown \\
    10 &electronics or technology & 11 &colorful or playful \\
    12 &outdoors & 13 &circular or round \\
    14 &paper-related or flat & 15 &sports-related or playing-related \\
    16 &tools or elongated & 17 &fluid-related or drink-related \\
    18 &water-related & 19 &oriented or many things \\
    20 &decay-related or grainy & 21 &white \\
    22 &coarse pattern or many things & 23 &red \\
    24 &long or thin & 25 &weapon-related or danger-related \\
    26 &black & 27 &household \\
    28 &feminine (stereotypical) & 29 &body part-related \\
    30 &tubular & 31 &music-related, hearing-related, or hobby-related \\
    32 &grid-related or grating-related & 33 &repetitive or spiky \\
    34 &construction-related or craftsman-related & 35 &spherical or voluminous \\
    36 &string-related or stringy & 37 &seating-related, standing-related, or lying-related \\
    38 &flying-related or sky-related & 39 &disgusting or slimy \\
    40 &elliptical or curved & 41 &sand-colored \\
    42 &green & 43 &bathroom-related or wetness-related \\
    44 &yellow & 45 &heat-related or light-related \\
    46 &beams-related or mesh-related & 47 &foot-related or walking-related \\
    48 &box-related or container & 49 &stick-shaped or cylindrical \\
    50 &head-related & 51 &upright, elongated, or voluminous \\
    52 &pointed or spiky & 53 &child-related or cute \\
    54 &farm-related or historical & 55 &seeing-related, small, or round \\
    56 &medicine-related & 57 &dessert-related \\
    58 &orange & 59 &thin or flat \\
    60 &cylindrical, conical, or cushioning & 61 &coldness-related or winter-related \\
    62 &measurement-related or numbers-related & 63 &fluffy or soft \\
    64 &masculine (stereotypical) & 65 &fine-grained pattern \\
    \bottomrule
  \end{tabular}
\end{table}

\newpage

\section{Prompts}
\label{apx:prompt-details}
In this section, we provide the full prompts used in our methods.

\subsection{Dimension-Guided Reasoning Enhancement Prompt}
\begin{tcolorbox}[
    breakable,
    colframe=gray!75, colback=black!2,
    width=\textwidth
]
\small
\textbf{\large Template} 

You're a knowledgeable expert. Given three objects \textbf{[triplet names]}. Here are their brief descriptions below.

\textbf{[triplet captions]}

With the above descriptions, you understand three objects \textbf{[triplet names]} simply. Next, you need to incorporate the vast amount of knowledge you have to understand them further.

You need to think carefully and analyse in terms of \textbf{[dimension name]}. Next, you need to give your analysis.

Here are some criteria you need to follow:

1.You only need to give your analysis. You must keep words in 200 words. You don't need to output the analysis process and how you analyse.

2.You need to analyse \textbf{[triplet names]} based on your vast knowledge, known information and analytical perspective which is \textbf{[dimension name]}.

3.Your analysis must be objective, reasonable, fair and logical.

4.If there is not much relationship between the angle of analysis, \textbf{[dimension name]} and one or more of \textbf{[triplet names]}, please do not carry out a forced analysis because forcing the analysis in terms of \textbf{[dimension name]} is not permissible to ensure your analysis is reasonable.

5.You must strictly follow the below response template to output with your analysis.

Here is your response template, You have to follow the template output given below

Analysis: {your analysis}

\tcblower 

\textbf{\large Example} 

triplet names: strawberry, pepper mill, and flask.

triplet captions: strawberry: sweet fleshy red fruit; pepper mill: a mill for grinding pepper; flask: bottle that has a narrow neck.

dimension name: plant related.

\textbf{Output:}
Analysis: The strawberry is a fruit that develops from the flower of the strawberry plant (Fragaria $\times$ ananassa), which is a member of the rose family (Rosaceae). It is characterized by its sweet, fleshy, and red flesh, and it is widely cultivated for its edible fruit. The plant itself is a perennial herb with runners that allow it to spread and propagate.

The pepper mill, while not directly related to plants, is used to grind peppercorns, which are the dried berries of the Piper nigrum plant, a flowering vine in the family Piperaceae. Peppercorns are harvested when they are fully mature and then dried, turning them into the familiar black, white, or green peppercorns.

The flask, being a container, does not have a direct botanical connection. However, flasks are often used in the processing and storage of plant-derived substances, such as essential oils, tinctures, and extracts, which are commonly used in various industries including pharmaceuticals, cosmetics, and food.

In summary, the strawberry is directly derived from a plant, the pepper mill is indirectly related through its use with peppercorns, and the flask can be associated with the storage and processing of plant materials.
\end{tcolorbox}
{
\captionsetup[table]{name=Prompt,hypcap=false}
\captionof{table}{Prompt for generating enhanced dimension-guided reasoning on triplet-level using LLM.}
\label{prompt:rationale-generation}
}

\subsection{O1O Task Training Prompt}
\begin{tcolorbox}[
    breakable,
    colframe=gray!75, colback=black!2,
    width=\textwidth
]
\small

\textbf{\large Template} 

\textbf{System:}

Here are three important notes:

1.Each image corresponds to a single object.

2.Focus only on the object itself and ignore any background elements.

3.Use reasonable reasoning and answer the question by thinking step by step.

Answer the following question:

\textbf{User:}

[triplet visual images]

Identify the object that is most different from the rest.

Options:

(A) Object A

(B) Object B

(C) Object C

Let's analyze each object.

\textbf{Assistant:}

\textbf{Object A}: [object A caption]

\textbf{Object B}: [object B caption]

\textbf{Object C}: [object C caption]

[dimension-guided reasoning]

It is clear that the most different object is [odd-one-out]. The answer is [right option].

\tcblower 

\textbf{\large Example}

\textbf{System:}

Here are three important notes:

1.Each image corresponds to a single object.

2.Focus only on the object itself and ignore any background elements.

3.Use reasonable reasoning and answer the question by thinking step by step.

Answer the following question:

\textbf{User:}
\medskip
\includegraphics[width=8cm]{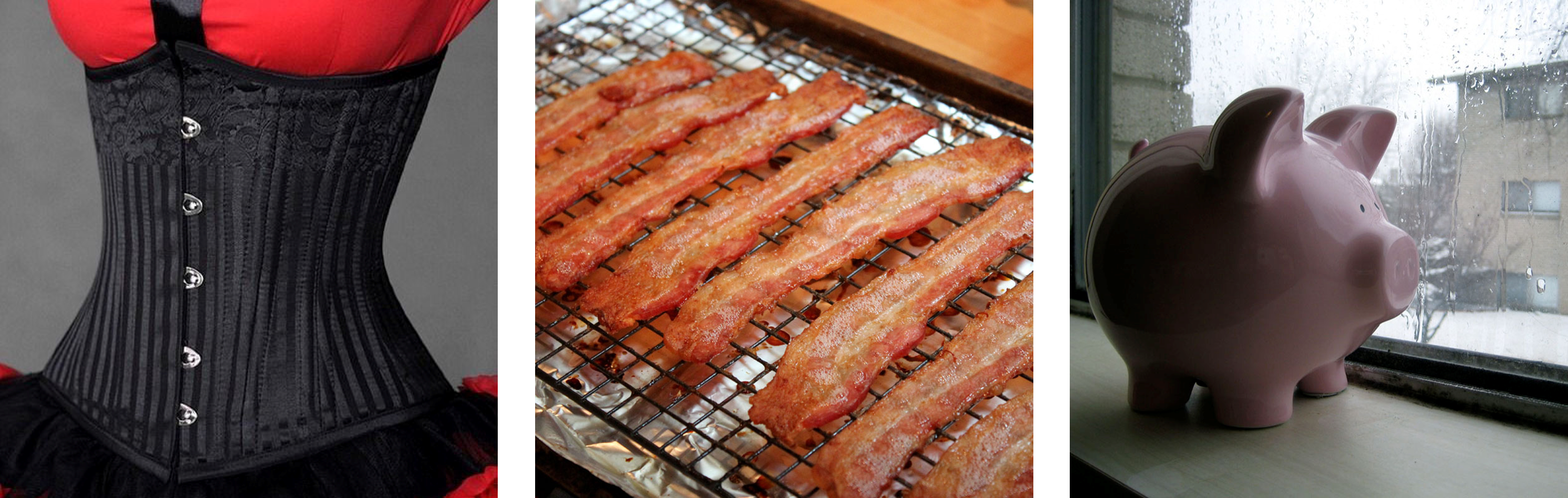}
\medskip

Identify the object that is most different from the rest.

Options:

(A) Object A

(B) Object B

(C) Object C

Let's analyze each object.

\textbf{Assistant:}

\textbf{Object A}: corset which is a closely fitted, typically sleeveless garment worn by women to provide support and shape to the torso, often extending from the bust to the hips, and usually fastened with laces or other closures.

\textbf{Object B}: bacon which is a cured meat from the back and sides of a hog, typically salted and dried or smoked, then sliced thin and often fried until crispy, serving as a popular breakfast food or ingredient in various dishes.

\textbf{Object C}: piggy bank which is a traditional children's savings container, typically made of ceramic or plastic, shaped like a pig, used to store coins and teach kids about saving money, often featuring a slot on top for depositing coins and a removable plug for withdrawing the accumulated savings.

Bacon, as a food item, is a cured meat product derived from the belly or side cuts of pork, often smoked and fried, serving as a staple in many cuisines. It is rich in flavor and fat, making it a popular ingredient in various dishes. The piggy bank, while not directly related to food, can symbolically represent savings that might be used to purchase food items, including bacon. However, it has no intrinsic connection to culinary practices or nutritional value. The corset, being a type of clothing, has no direct relationship to food. Historically, corsets were designed for shaping the body and have no functional or symbolic link to food preparation, consumption, or culture. Therefore, while bacon is inherently food-related, the piggy bank and corset do not have significant connections to the food domain.

It is clear that the most different object is \textbf{corset}. The answer is A.
\end{tcolorbox}
{
\captionsetup[table]{name=Prompt,hypcap=false}
\captionof{table}{Prompt for O1O task training data to align the latent structure of human cognition.}
\label{prompt:o1o-qa}
}

\subsection{O1O Direct Inference Prompt}
\begin{tcolorbox}[
    breakable,
    colframe=gray!75, colback=black!2,
    width=\textwidth
]
\small

\textbf{\large Input} 

\textbf{System:}

Here are three important notes:

1.Each image corresponds to a single object.

2.Focus only on the object itself and ignore any background elements.

3.Answer the question with correct option directly (e.g., A, B, or C).

Answer the following question:

\textbf{User:}

[triplet visual images]

Which object is the odd one out?

Options:

(A) Object A

(B) Object B

(C) Object C

\end{tcolorbox}
{
\captionsetup[table]{name=Prompt,hypcap=false}
\captionof{table}{Prompt for O1O direct inference. MLLMs only answer with predicted option (A/B/C).}
\label{prompt:o1o-direct}
}

\subsection{O1O CoT Inference Prompt}
\begin{tcolorbox}[
    breakable,
    colframe=gray!75, colback=black!2,
    width=\textwidth
]
\small

\textbf{\large Standard CoT} 

\textbf{System:}

Here are three important notes:

1.Each image corresponds to a single object.

2.Focus only on the object itself and ignore any background elements.

3.Use reasonable reasoning and answer the question by thinking step by step.

Answer the following question:

\textbf{User:}

[triplet visual images]

Which object is the odd one out?

Options:

(A) Object A

(B) Object B

(C) Object C

At the very last line of your response, must put correct answer: {your chosen option}. Do not add anything after it.

Let's analyze each object.
\end{tcolorbox}
{
\captionsetup[table]{name=Prompt,hypcap=false}
\captionof{table}{Prompt for O1O standard CoT inference. MLLMs need to think step-by-step and give final choice. This prompt will be used by baseline and CogAlign for CoT, S-CoT metrics.}
\label{prompt:o1o-cot-infer}
}

\par\medskip

\begin{tcolorbox}[
    breakable,
    colframe=gray!75, colback=black!2,
    width=\textwidth
]
\small

\textbf{\large $d^*$ CoT} 

\textbf{System:}

Here are two important notes:

1.Each image corresponds to a single object.

2.Focus only on the object itself and ignore any background elements.

Answer the following question:

\textbf{User:}

[triplet visual images]

Which object is the odd one out?

Options:

(A) Object A

(B) Object B

(C) Object C.

\textbf{Thinking Perspective:}

Before providing your final choice, you must conduct deep, step by step thinking from the perspective of \textbf{[dimension name]}.

\textbf{CRITICAL Output Format:}

You MUST strictly adhere to the exact structure below. Do not add any extra words outside of this template.

<thinking>

[Write your detailed step-by-step reasoning process here]

</thinking>

Therefore, the answer is <choice>[A, B, or C]</choice>
\end{tcolorbox}
{
\captionsetup[table]{name=Prompt,hypcap=false}
\captionof{table}{Prompt for O1O $d^*$ dimension-guided CoT inference. MLLMs need to think step-by-step and give final choice under the guidance of $d^*$ semantic dimension. This prompt will be used by baseline models for $d^*$-CoT metric.}
\label{prompt:o1o-cot-infer-dcot}
}

\par\medskip

\begin{tcolorbox}[
    breakable,
    colframe=gray!75, colback=black!2,
    width=\textwidth
]
\small

\textbf{\large FS CoT} 

\textbf{System:}

Here are two important notes:

1.Each image corresponds to a single object.

2.Focus only on the object itself and ignore any background elements.

Answer the following question:

\textbf{User:}

[triplet visual images]

Which object is the odd one out?

Options:

(A) Object A

(B) Object B

(C) Object C.

\textbf{Available Semantic Dimensions:}

dim1: metallic or artificial

dim2: food-related

dim3: animal-related

dim4: textile

dim5: plant-related

dim6: house-related or furnishing-related

dim7: valuable or precious

dim8: transportation-related or movement-related

dim9: body-related or people-related

dim10: wood-related or brown

dim11: electronics or technology

dim12: colorful or playful

dim13: outdoors

dim14: circular or round

dim15: paper-related or flat

dim16: sports-related or playing-related

dim17: tools or elongated

dim18: fluid-related or drink-related

dim19: water-related

dim20: oriented or many things

dim21: decay-related or grainy

dim22: white

dim23: coarse pattern or many things

dim24: red

dim25: long or thin

dim26: weapon-related or danger-related

dim27: black

dim28: household

dim29: feminine (stereotypical)

dim30: body part-related

dim31: tubular

dim32: music-related, hearing-related, or hobby-related

dim33: grid-related or grating-related

dim34: repetitive or spiky

dim35: construction-related or craftsman-related

dim36: spherical or voluminous

dim37: string-related or stringy

dim38: seating-related, standing-related, or lying-related

dim39: flying-related or sky-related

dim40: disgusting or slimy

dim41: elliptical or curved

dim42: sand-colored

dim43: green

dim44: bathroom-related or wetness-related

dim45: yellow

dim46: heat-related or light-related

dim47: beams-related or mesh-related

dim48: foot-related or walking-related

dim49: box-related or container

dim50: stick-shaped or cylindrical

dim51: head-related

dim52: upright, elongated, or voluminous

dim53: pointed or spiky

dim54: child-related or cute

dim55: farm-related or historical

dim56: seeing-related, small, or round

dim57: medicine-related

dim58: dessert-related

dim59: orange

dim60: thin or flat

dim61: cylindrical, conical, or cushioning

dim62: coldness-related or winter-related

dim63: measurement-related or numbers-related

dim64: fluffy or soft

dim65: masculine (stereotypical)

dim66: fine-grained pattern

\textbf{Thinking Perspective:}

Before providing your final choice, you must conduct deep, step by step thinking from the perspective of one or multiple dimensions chosen from the \textbf{Available Semantic Dimensions} list above.

\textbf{CRITICAL Output Format:}

You MUST strictly adhere to the exact structure below. Do not add any extra words outside of this template.

<thinking>

[Write your detailed step-by-step reasoning process here]

</thinking>

Therefore, the answer is <choice>[A, B, or C]</choice>

\end{tcolorbox}
{
\captionsetup[table]{name=Prompt,hypcap=false}
\captionof{table}{Prompt for O1O 66 dimensions CoT inference. MLLMs need to think step-by-step from one or more semantic dimension(s) and give final choice. This prompt will be used by baseline models for FS-CoT metric.}
\label{prompt:o1o-cot-infer-fscot}
}

\section{Impact Statement}
\label{sec:impact}
This work presents research on aligning Multimodal Large Language Models (MLLMs) with latent structure of human cognition. The primary positive impact is advancing our understanding of MLLMs by thinking process before decision-making, providing a data-driven framework to bridge the representational gap between models and human similarity judgements, making MLLMs more human-like. However, because our framework relies on crowdsourced behavioral data provided by Hebart et al., it carries the inherent risk of internalizing and transferring implicit human cognitive biases or stereotypes present in the dataset. To mitigate this, we advise future research to audit behavior-derived datasets for cultural and demographic biases prior to alignment.

\section{Dataset Licenses}
\label{sec:dataset-licenses}
We use the M4-Instruct dataset \citep{li2024llava}, THINGS behavioral dataset \citep{hebart2020revealing,hebart2023things}, and Natural Scenes Dataset (NSD) \citep{allen2022massive} following their respective license agreements. Both datasets are accessed through their official releases and use exclusively for academic research purposes.


\newpage
\section*{NeurIPS Paper Checklist}

\begin{enumerate}
  \item {\bf Claims}
  \item[] Question: Do the main claims made in the abstract and introduction accurately reflect the paper's contributions and scope?
  \item[] Answer: \answerYes{} 
  \item[] Justification: We summarize the main contributions in the Introduction section~\ref{sec:intro}.

  \item {\bf Limitations}
  \item[] Question: Does the paper discuss the limitations of the work performed by the authors?
  \item[] Answer: \answerYes{} 
  \item[] Justification: We discuss the limitations in the Limitations section of the Section~\ref{sec:limitions}.

  \item {\bf Theory assumptions and proofs}
  \item[] Question: For each theoretical result, does the paper provide the full set of assumptions and a complete (and correct) proof?
  \item[] Answer: \answerNA{} 
  \item[] Justification: \answerNA{}

  \item {\bf Experimental result reproducibility}
  \item[] Question: Does the paper fully disclose all the information needed to reproduce the main experimental results of the paper to the extent that it affects the main claims and/or conclusions of the paper (regardless of whether the code and data are provided or not)?
  \item[] Answer: \answerYes{} 
  \item[] Justification: We disclose data preparation, training and evaluation settings, feature extraction, and prompts in Section~\ref{sec:experiments} and Appendix~\ref{apx:feature-extraction-method}, ~\ref{apx:data-details}, ~\ref{apx:experimental-details}, and ~\ref{apx:prompt-details}. The released code and rationale data provide the configurations and procedures.

  \item {\bf Open access to data and code}
  \item[] Question: Does the paper provide open access to the data and code, with sufficient instructions to faithfully reproduce the main experimental results, as described in supplemental material?
  \item[] Answer: \answerYes{} 
  \item[] Justification: The code and data is available at \url{https://github.com/ening-open/cogalign}.

  \item {\bf Experimental setting/details}
  \item[] Question: Does the paper specify all the training and test details (e.g., data splits, hyperparameters, how they were chosen, type of optimizer) necessary to understand the results?
  \item[] Answer: \answerYes{} 
  \item[] Justification: We specify the training and test details in the Section~\ref{sec:experiments} and Appendix~\ref{apx:data-details}, and ~\ref{apx:experimental-details}.

  \item {\bf Experiment statistical significance}
  \item[] Question: Does the paper report error bars suitably and correctly defined or other appropriate information about the statistical significance of the experiments?
  \item[] Answer: \answerYes{} 
  \item[] Justification: We provide confidence intervals and statistical significance tests in the experiments.

  \item {\bf Experiments compute resources}
  \item[] Question: For each experiment, does the paper provide sufficient information on the computer resources (type of compute workers, memory, time of execution) needed to reproduce the experiments?
  \item[] Answer: \answerYes{} 
  \item[] Justification: We provide detailed training and evaluation settings in Appendix~\ref{apx:experimental-details}.

  \item {\bf Code of ethics}
  \item[] Question: Does the research conducted in the paper conform, in every respect, with the NeurIPS Code of Ethics \url{https://neurips.cc/public/EthicsGuidelines}?
  \item[] Answer: \answerYes{} 
  \item[] Justification: This research uses publicly available behavioral datasets and fMRI datasets collated under the original studies' ethical approvals.

  \item {\bf Broader impacts}
  \item[] Question: Does the paper discuss both potential positive societal impacts and negative societal impacts of the work performed?
  \item[] Answer: \answerYes{} 
  \item[] Justification: We discuss broader impacts in Appendix~\ref{sec:impact}.

  \item {\bf Safeguards}
  \item[] Question: Does the paper describe safeguards that have been put in place for responsible release of data or models that have a high risk for misuse (e.g., pre-trained language models, image generators, or scraped datasets)?
  \item[] Answer: \answerNA{} 
  \item[] Justification: \answerNA{}

  \item {\bf Licenses for existing assets}
  \item[] Question: Are the creators or original owners of assets (e.g., code, data, models), used in the paper, properly credited and are the license and terms of use explicitly mentioned and properly respected?
  \item[] Answer: \answerYes{} 
  \item[] Justification: We have cited all necessary papers and the license information and usage details of the datasets are provided in Appendix~\ref{sec:dataset-licenses}.

  \item {\bf New assets}
  \item[] Question: Are new assets introduced in the paper well documented and is the documentation provided alongside the assets?
  \item[] Answer: \answerYes{} 
  \item[] Justification: The released code and related data are documented at \url{https://github.com/ening-open/cogalign}.

  \item {\bf Crowdsourcing and research with human subjects}
  \item[] Question: For crowdsourcing experiments and research with human subjects, does the paper include the full text of instructions given to participants and screenshots, if applicable, as well as details about compensation (if any)?
  \item[] Answer: \answerNA{} 
  \item[] Justification: This work does not involve crowdsourcing experiments or research with human subjects.

  \item {\bf Institutional review board (IRB) approvals or equivalent for research with human subjects}
  \item[] Question: Does the paper describe potential risks incurred by study participants, whether such risks were disclosed to the subjects, and whether Institutional Review Board (IRB) approvals (or an equivalent approval/review based on the requirements of your country or institution) were obtained?
  \item[] Answer: \answerNA{} 
  \item[] Justification: \answerNA{}

  \item {\bf Declaration of LLM usage}
  \item[] Question: Does the paper describe the usage of LLMs if it is an important, original, or non-standard component of the core methods in this research? Note that if the LLM is used only for writing, editing, or formatting purposes and does \emph{not} impact the core methodology, scientific rigor, or originality of the research, declaration is not required.
  \item[] Answer: \answerYes{} 
  \item[] Justification: LLMs are employed to convert the most salient decision dimension into a dimension-grounded linguistic rationale, which may impact the alignment performance.
\end{enumerate}

\end{document}